\documentclass[runningheads]{llncs}

\usepackage{eccv}

\usepackage{eccvabbrv}

\usepackage{graphicx}
\usepackage{booktabs}

\usepackage[accsupp]{axessibility}  

\usepackage{hyperref}

\usepackage{orcidlink}

\usepackage[table]{xcolor}
\usepackage{caption}
\usepackage{subcaption}
\usepackage{booktabs,multirow,arydshln}
\usepackage{amsmath,amssymb} 
\usepackage{titletoc}

\definecolor{gold}{HTML}{FFF9E5}
\definecolor{cornellred}{rgb}{0.7, 0.11, 0.11}
\definecolor{cadmiumgreen}{rgb}{0.0, 0.42, 0.24}
\definecolor{Blue9}{rgb}{0.098,0.3,0.9}

\hypersetup{
    colorlinks=true,
    linkcolor=cornellred,   
    urlcolor=Blue9,
    citecolor=cadmiumgreen
}

\begin{document}

\title{Open-Vocabulary BEV Segmentation with 3D-Aware Geometric Constraints}

\titlerunning{OVBEVSeg}

\author{Hojun Choi\inst{1}$^*$\orcidlink{0009-0009-4121-4519}\and
Seulbin Hwang\inst{2}\orcidlink{0009-0009-3448-1862} \and
Daejung Kim\inst{2}\orcidlink{0000-0002-5517-1179} \and
Kisung Kim\inst{2}\orcidlink{0009-0006-8729-6458} \and\\
Hyunjung Shim\inst{1}$^\dagger$\orcidlink{0000-0001-6796-1058} \and
Jinhan Lee\inst{2}$^\dagger$\orcidlink{0009-0003-2965-143X}}

\authorrunning{H.~Choi et al.}

\institute{KAIST AI, South Korea\\
\email{\{hchoi256,kateshim\}@kaist.ac.kr} \and
NAVER LABS, South Korea\\
\email{\{h.sb,daejung.kim,ks.kim,jinhan.lee\}@naverlabs.com}}

\maketitle

\begingroup
\renewcommand{\thefootnote}{}
\footnotetext{Project page: \url{https://hchoi256.github.io/projects/ovbevseg}.}
\footnotetext{This paper has been accepted to ECCV 2026.}
\footnotetext{$^*$Work done during an internship at NAVER LABS.}
\footnotetext{$^\dagger$Co-corresponding authors}
\endgroup

\begin{abstract}
  Bird's-eye view (BEV) perception fuses multi-camera images into a unified top-down representation for autonomous driving. Despite recent progress, state-of-the-art methods remain confined to closed-set scenarios, making them vulnerable to unpredictable real-world environments. In this work, we introduce open-vocabulary BEV segmentation (OVBS), which leverages vision-language models (VLMs) to recognize categories beyond the training set while maintaining precise BEV perception and real-time efficiency. A key challenge in OVBS lies in the 3D geometric inconsistency inherent in the ill-posed lifting of 2D VLM semantics into BEV. To address this, we propose OVBEVSeg, a geometry-aware OVBS framework that enhances efficient Gaussian splatting (GS)-based unprojection by leveraging robust 3D geometric constraints across three progressive stages: (1) 2D-to-BEV pseudo-labeling via reliable 3D projection for OV generalization; (2) joint 2D–BEV per-scene optimization with BEV structural constraints for 3D geometric consistency; and (3) 3D geometric distillation for online efficiency. On the nuScenes dataset, OVBEVSeg achieves state-of-the-art performance, outperforming closed-set methods by 15.3 mIoU on unseen categories. Remarkably, even with no novel-class ground-truth labels, it remains competitive with self- and semi-supervised baselines trained with up to 40\% of ground-truth annotations. Furthermore, it achieves 2.5$\times$ faster inference with only 0.22$\times$ the memory consumption of projection-based methods.
\end{abstract}


\section{Introduction}
\label{sec:intro}

Bird's-eye view (BEV) perception has emerged as a foundational paradigm in autonomous driving (AD), providing a unified ego-centric top-down representation for effective multi-sensor fusion. It now underpins a broad spectrum of 3D scene understanding tasks, including object detection~\cite{bevformer,petrv2,m2bev}, semantic segmentation~\cite{pointbev,simplebev,fiery,lss,cvt}, and trajectory planning~\cite{fiery,uniad,stp3}. A central challenge in BEV perception is how to establish spatial correspondence between 2D camera features and the BEV representation. This process is fueled by advancements in depth-based~\cite{fiery,bevdet,lss,bevdepth}, projection-based~\cite{pointbev,simplebev}, attention-based~\cite{bevformer,cvt}, and Gaussian-based~\cite{gaussianbev,gaussianlss} architectures, leading to successful real-time, scalable inference BEV perception critical for modern autonomous platforms.

\begin{figure}[tb]
  \centering
  \includegraphics[width=\linewidth]{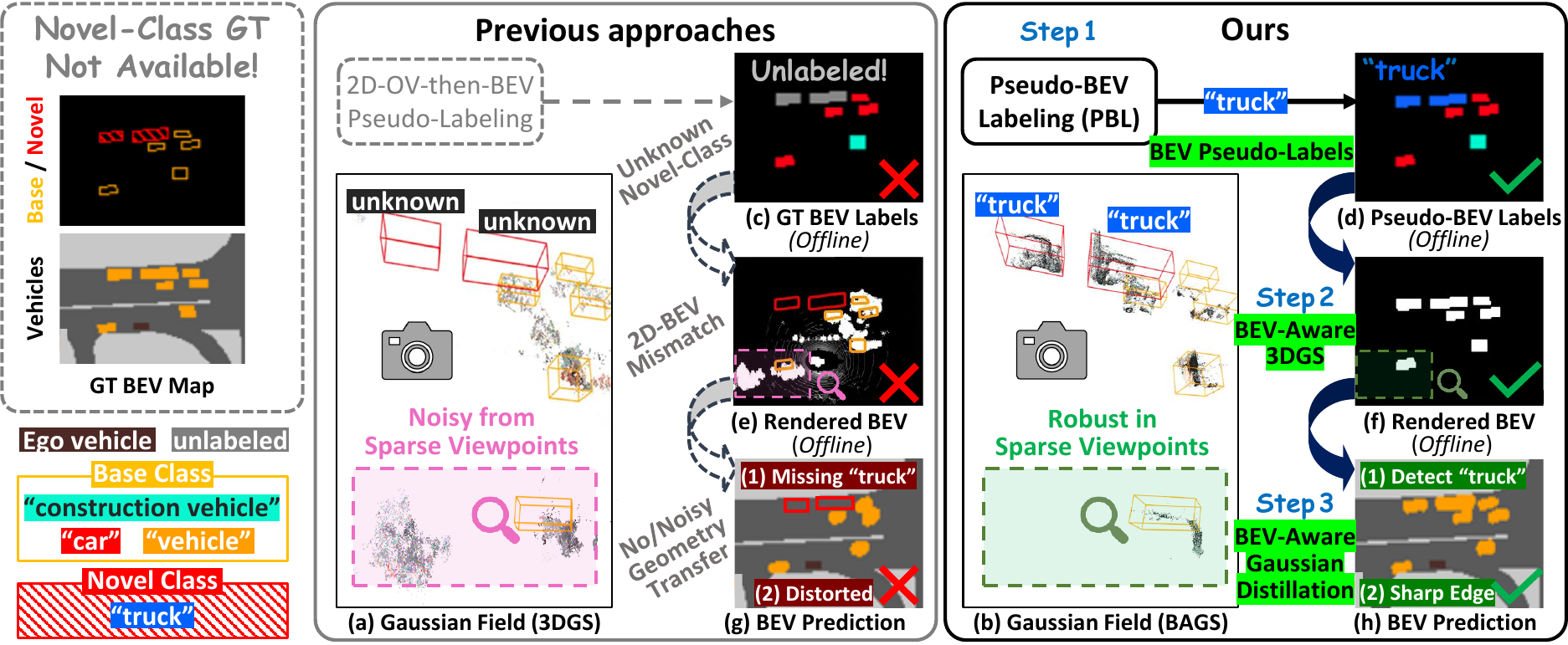}
  \caption{\textbf{(c,d)} Ground-truth or pseudo BEV annotations. \textbf{(a,e)} 3D and BEV renderings with plain 3D Gaussian splatting, showing dispersed splats beyond object regions. \textbf{(b,f)} 3D and BEV renderings with our BAGS, with splats concentrated along object boundaries. \textbf{(g,h)} Online BEV predictions: the baseline misses novel classes (\eg, ``truck'') and exhibits boundary distortions, which are successfully resolved by our method.}
  \label{fig:1_limit}
\end{figure}

Despite strict demands for efficient deployment, a crucial yet overlooked assumption persists: existing BEV perception pipelines are fundamentally framed as closed-set systems. That is, both training and inference assume the same, finite set of semantic classes. In real-world AD, this assumption is neither realistic nor safe. In~\cref{fig:1_limit}(g) and~\ref{fig:4_oval}, autonomous vehicles continuously encounter previously unseen objects—\eg, ``truck,'' or atypical movers like ``stroller'' or ``wheelchair.'' These novel classes, while present in sensor data, lack both human-annotated and ground-truth labels during training and are therefore silently ignored during inference, creating semantically blind regions in the BEV map. This discrepancy is more than a technical shortcoming: it poses a direct threat to downstream tasks such as path planning and collision avoidance~\cite{uniad,driveadapter}. We argue that reframing BEV perception for open-world environments is not optional but essential for ensuring semantic completeness and safety in real-world autonomy.

To this end, we introduce \textit{open-vocabulary BEV segmentation} (OVBS), a new task that extends OV recognition from 2D vision to the BEV domain, leveraging vision-language models (VLMs) to recognize previously unseen categories. The objective of OVBS is to simultaneously achieve three goals: OV generalization to novel classes, precise BEV perception, and inference-time efficiency.

A natural extension to OVBS is to graft OV visual recognition~\cite{ov,ovsurvey} onto the core logic of unprojection-based BEV pipelines. Concretely, one could attach a 2D OV detector~\cite{groundingdino} to each camera view, lift the predicted 2D regions into 3D space via depth estimation, and project them into BEV—often used as a fair OV baseline. However, this seemingly straightforward ``2D-OV-then-BEV'' solution remains vulnerable to a longstanding challenge in AD: it relies on direct 2D-to-3D lifting, which is fundamentally brittle in large-scale outdoor scenes with sparse views. Small errors in 2D localization or semantic classification are amplified by the ill-posed lifting process~\cite{lss,bevdepth}—a structural flaw inherent to all unprojection-based methods. In~\cref{fig:1_limit}, this direct mapping fails on two fronts: \textbf{(L1)} (c) compounded errors hinder the recognition of novel classes in BEV; and \textbf{(L2)} (g) inherent geometric instability results in geometrically inconsistent or distorted BEV boxes. In short, naive designs built upon 2D unprojection break down due to the lack of consistent multi-view 3D geometry—a prerequisite for both OV-to-BEV extensions and reliable unprojection-based BEV perception.

This inherent vulnerability persists even in Gaussian splatting (GS)-based unprojection—arguably the most promising candidate due to its superior efficiency and scalability—as it remains susceptible to the same 2D-to-3D lifting errors (L1, L2). Specifically, it amplifies these biases by directly lifting 2D features into 3D Gaussians for BEV splatting. To address these limitations, our key insight is that the direction of geometric reasoning must be reversed: rather than lifting noisy 2D predictions into 3D (unprojection), we should project reliable 3D structure down into 2D and BEV (projection). Following this projective shift, we propose OVBEVSeg, a novel geometry-aware OVBS framework that leverages robust 3D geometric constraints across three progressive stages: (1) generating BEV pseudo-labels by establishing consistent 2D–BEV correspondences and routing 2D VLM semantics via reliable 3D projection of class-agnostic 3D detections~\cite{union}, thereby bypassing erroneous 2D unprojection and resolving (L1); (2) conducting joint 2D–BEV per-scene optimization constrained by BEV spatial layouts as strong geometric constraints, thereby enabling robust 3D geometric reconstruction from sparse viewpoints; and (3) distilling the recovered geometry into a student model, mitigating (L2) while preserving inference efficiency. Crucially, these stages are not independent modules but successive instantiations of a single principle: leveraging 3D geometric constraints to bypass the ill-posed 2D unprojection—first for semantics, geometry, and efficient knowledge transfer.

Extensive experiments show that OVBEVSeg sets a new state of the art, surpassing prior work~\cite{pointbev} by 15.3 mIoU on novel classes while running 2.5$\times$ faster with 0.22$\times$ memory usage. Without novel-class ground-truth labels, it remains comparable to semi-/self-supervised baselines using up to 40\% annotations.


\section{Related Work}
\label{sec:related}

\subsubsection{Bird's-Eye View (BEV) Perception.}
BEV perception maps multi-sensor data into a unified top-down space for 3D scene understanding. Existing methods fall into four paradigms: (1) depth-based methods~\cite{bevdet,bevdepth,lss} explicitly splat image features with per-pixel depth distributions into a 3D grid, yielding geometrically grounded BEV; (2) projection-based schemes~\cite{pointbev,simplebev} query camera features at predefined 3D points; (3) attention-based transformers~\cite{bevformer,sparsebev,petrv2,wang,cvt} leverage cross-attention to fuse image and BEV tokens; and (4) feed-forward 3D Gaussian splatting methods~\cite{gaussianbev,gaussianlss} predict 3D Gaussians to rasterize BEV features. In particular, depth-based approaches are highly sensitive to depth estimation errors, which can propagate into the BEV representation. Despite their architectural diversity, all these paradigms remain strictly constrained to closed-set categories. In this work, we transcend these boundaries by recasting BEV perception as an open-vocabulary task integrated with 3D geometric constraints.

\subsubsection{3D Gaussian Splatting (3DGS).}
3DGS~\cite{3dgs} has emerged as a powerful differentiable renderer for scene representation. In autonomous driving, 3DGS-based methods bifurcate into \textit{offline} per-scene optimization and \textit{online} feed-forward approaches. Within this taxonomy, offline methods~\cite{drivinggaussian,hugs} incrementally optimize static Gaussian fields while compositing object-level dynamics. While recovering high-fidelity geometry under photometric supervision, they remain computationally prohibitive for real-time perception. Conversely, feed-forward systems~\cite{pixelsplat,mvsplat,depthsplat} enable omnidirectional rendering by computing per-pixel ray directions under equirectangular models. Building on this, recent BEV variants~\cite{gaussianlss,gaussianbev} map image features to 3D Gaussians for BEV splatting. Despite their scalability, these methods often compromise 3D geometric precision due to the ill-posed nature of 2D-to-3D unprojection under sparse viewpoints. Drawing on hybrid 3DGS advances~\cite{ufo,instantsplat}, we bridge this gap by synergizing the geometric rigor of per-scene optimization with feed-forward efficiency.

\subsubsection{Open-Vocabulary Learning (OVL).}
To overcome the scalability limits of fixed categories, OVL~\cite{ov} has gained traction, leveraging large-scale vision-language pretraining~\cite{clip} for unseen category recognition. In autonomous driving, several studies have extended OVL to various 3D tasks: OV-3DETIC~\cite{ov3detic} addresses novel-class 3D detection via image-level supervision and debiased cross-modal contrastive learning; OpenScene~\cite{openscene} enables OV semantic segmentation (OVSS) by co-embedding 3D features with text and pixels; and OVTrack~\cite{ovtrack} introduces the first OV multi-object tracking (OV-MOT) framework by distilling vision-language knowledge into a tracker; and LidarCLIP~\cite{lidarclip} aligns LiDAR features with 2D embeddings for OV recognition in point clouds. In contrast, OV recognition within the BEV space remains underexplored, hindered by the 2D--BEV correspondence ambiguity. To our knowledge, this is the first \textit{open-vocabulary BEV segmentation} (OVBS) framework that leverages robust 3D geometric constraints to map 2D VLM semantics consistently into the BEV space.


\section{Methodology}
\label{sec:method}

In this work, we introduce \textit{open-vocabulary BEV segmentation} (OVBS), a novel task aimed at enhancing generalization to unseen categories via vision-language models (VLMs) while preserving precise BEV perception and real-time efficiency. As established in~\cref{sec:intro}, the core bottleneck for OVBS can be the ill-posed 2D-to-3D lifting shared by all unprojection-based BEV methods. Our 3D projection-first principle addresses this by reversing the direction of geometric reasoning: we project reliable 3D structure down into 2D and BEV.

In~\cref{fig:2_main}, we propose OVBEVSeg, a geometry-aware OVBS framework that enhances Gaussian splatting (GS)-based unprojection by incorporating robust 3D geometric constraints across three synergistic modules: (1) pseudo-BEV labeling (PBL) for OV generalization (\cref{sec:method:c1}), (2) BEV-aware 3DGS (BAGS) for geometric reconstruction (\cref{sec:method:c2}), and (3) BEV-aware Gaussian distillation (BAGD) for efficiency (\cref{sec:method:c3}).
In (1), we explicitly establish 2D–BEV correspondence by propagating semantics via the 3D projection of unsupervised 3D clusters~\cite{union}. These 3D candidates are projected into the image space, where instance masks are recovered using a segmentation foundation model~\cite{groundedsam}. For every segmented 2D object, we extract CLIP-based image embeddings and obtain pseudo-class labels via text prompts. These labels are then mapped back to corresponding 3D bounding boxes, which serve as semantic carriers that link the 2D image domain to the 3D scene layout. By projecting these annotated 3D boxes into the BEV space, we generate consistent BEV pseudo-labels for both base (\eg, ``car'') and novel classes (\eg, ``truck''); see~\cref{fig:1_limit}(d). Building upon these correspondences, in (2), BAGS leverages the derived BEV structural layouts as strong geometric priors to enforce 3D geometric consistency from sparse viewpoints; see~\cref{fig:1_limit}(b,f). In (3), BAGD distills this high-fidelity geometric knowledge into a student network to ensure real-time inference; see~\cref{fig:1_limit}(h). These stages successively leverage 3D geometric constraints to bypass 2D unprojection across semantics, geometry, and knowledge transfer, respectively.

\subsection{Open-Vocabulary Pseudo-BEV Labeling (PBL)}
\label{sec:method:c1}
In this section, we rethink standard BEV perception through the lens of OVBS, targeting robust novel-class recognition under base-class supervision alone. Unlike naive 2D-to-3D lifting, our 3D projection-first principle ensures (1) consistent 2D–BEV semantic propagation by leveraging unsupervised 3D detections as geometric constraints, thereby facilitating (2) BEV pseudo-label generation necessary for OV recognition—a linchpin of safety-critical AD.

\subsubsection{Generating 2D–BEV Correspondences via 3D Projection.}
A straightforward route to consistent 2D–BEV object discovery is to run an off-the-shelf OV 2D detector~\cite{groundingdino}, lift the detections to 3D via depth estimation~\cite{fiery,gaussianlss}, and then project them into BEV. This protocol is widely adopted in OV settings for fairness, since the detector is not trained on the dataset's novel-class ground truth~\cite{opengaussian,groundedsam,zeng}. However, these pipelines typically rely on ray sampling for depth and inevitably discard information in the 2D-to-3D transformation~\cite{lss,bevdepth}. Despite notable progress in 2D-to-3D understanding, lifting image features into a metrically consistent 3D space remains fundamentally ill-posed and approximate due to occlusions, depth ambiguity, and discretization.

\begin{figure}[tb]
  \centering
   \includegraphics[width=\linewidth]{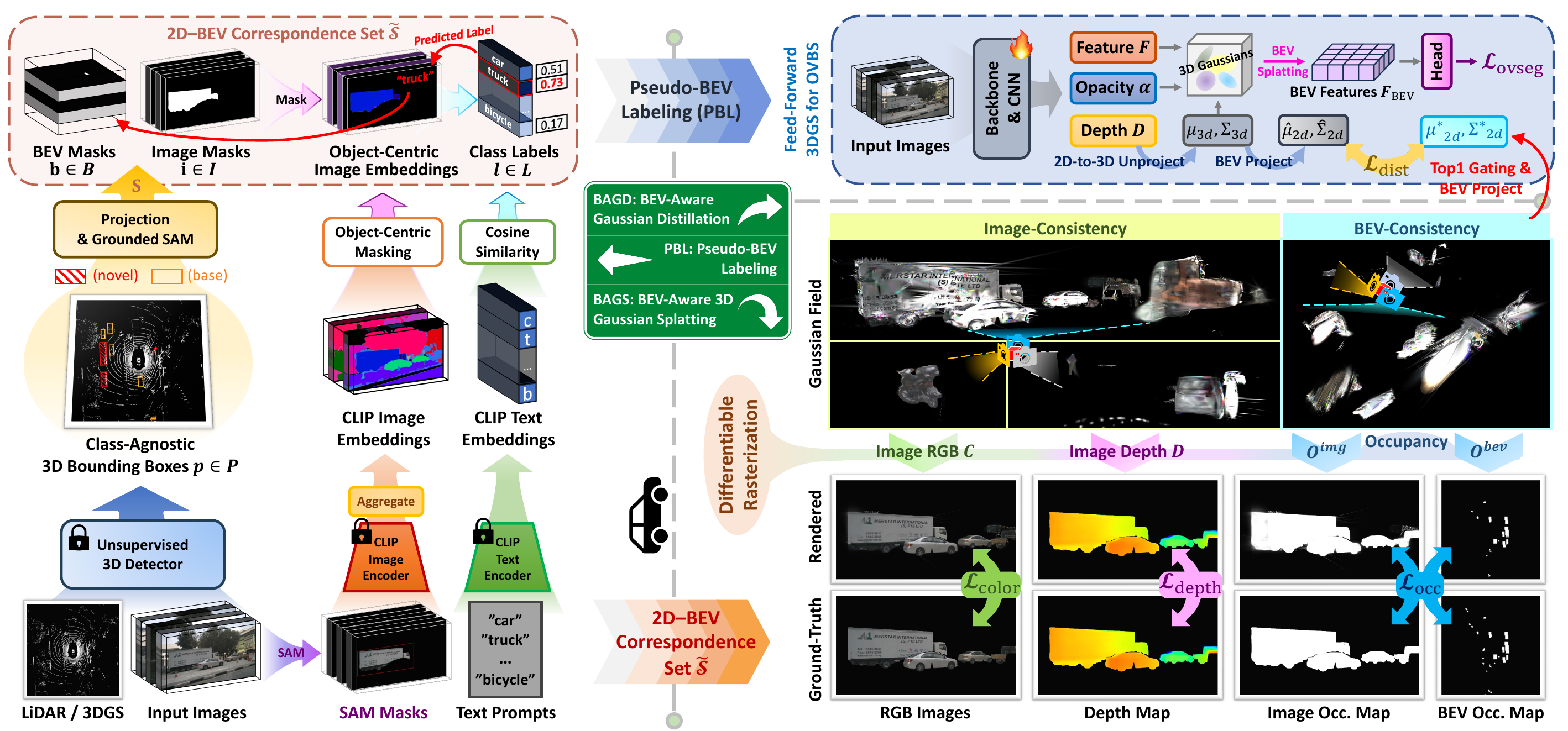}
   \caption{Our method consists of three modules: \textbf{(1) PBL} generates BEV pseudo-labels by leveraging unsupervised 3D object discovery~\cite{union}, foundational 2D segmentation~\cite{groundedsam}, and VLM grounding~\cite{clip} to construct a 2D--BEV correspondence set $\tilde{\mathcal{S}}$. \textbf{(2) BAGS} performs 3D geometry reconstruction using 2D, depth, and BEV occupancy supervision from $\tilde{\mathcal{S}}$. \textbf{(3) BAGD} distills this optimized 3D geometry into a feed-forward 3DGS-based student network in the BEV space, preserving fast inference speeds.}
   \label{fig:2_main}
\end{figure}

Instead, we avoid this unprojection ambiguity by predicting 3D object boxes with an unsupervised 3D detector and projecting them onto the image and BEV planes to establish reliable 2D–BEV correspondences. Specifically, we adopt the 3D bounding-box pseudo-annotations $P$ from UNION~\cite{union}, which deliver competitive class-agnostic detection by leveraging multi-sensor and temporal cues. Given these boxes, we invoke a 2D segmentation foundation model~\cite{groundedsam} to obtain pixel-level masks for dynamic objects within the projected pseudo-boxes. For objects seen by multiple cameras, we retain only the largest 2D instance $i \in I$ to maximize visual representation, enforcing a one-to-one match with a single BEV instance $b \in B$. Finally, we define the 2D--BEV correspondence set $\mathcal{S}=\{(i,b,p)\mid i\in I,\ b\in B,\ p\in P\}$. As shown in~\cref{fig:2_main}, it successfully encodes object-centric correspondences (\eg, ``truck'') across 2D and BEV domains. This 3D-driven pipeline effectively circumvents the unstable depth estimation inherent in naive 2D unprojection---a consistent cross-modal route for 2D VLM semantics.

To mitigate failure modes arising from noisy 3D detection priors, we discard instances where the 2D mask area significantly deviates from its projected 2D bounding box. This mismatch signals that the 3D box is inadequately sized; specifically, an over-estimated 3D box leads to a substantial area discrepancy, while an under-estimated box typically yields a non-object response from the foundational grounding model. Such proactive pruning ensures that the resulting set $\mathcal{S}$ remains high-fidelity, even when raw 3D predictions are suboptimal.

\subsubsection{Generating BEV Pseudo-Labels.}
Regarding OV pseudo-labeling, we adopt the standard CLIP-based paradigm, leveraging its shared embedding space to align visual and linguistic concepts. Specifically, for each 2D region proposal, the image encoder generates a visual embedding, which is then compared against a set of text embeddings---encompassing both base and novel classes---to compute cosine similarity scores. These textual prompts can be derived from image captions~\cite{vild,ovrcnn}, dataset class names~\cite{vlplm}, or MLLM-driven zero-shot labels~\cite{mspl}.

Following this paradigm, we initialize $C$ categories from the dataset~\cite{nuscenes} using their text embeddings derived from standard prompt templates~\cite{vild}. Using $\mathcal{S}$, we first extract 2D instance masks $I$ to generate background-masked, object-centric crops for the CLIP image encoder, thereby mitigating semantic ambiguity near object boundaries. Each visual embedding is then assigned a pseudo-label $l$ based on its maximum cosine similarity to the text embeddings of the $C$ categories. By incorporating these labels into the original set---$\tilde{\mathcal{S}}=\{(i,b,p,l)\mid(i, b, p)\in S,\ l \in L\}$---we associate the pseudo-labels with their respective 3D instances, effectively lifting image-level VLM outputs into the BEV space. In~\cref{fig:2_main}, VLM embeddings for novel classes (\eg, ``truck'') remain spatially consistent across both domains. As a result, PBL bridges the gap between closed-set training and OV generalization while relying strictly on base-class supervision.

\subsection{BEV-Aware 3D Gaussian Splatting (BAGS)}
\label{sec:method:c2}
In this section, we directly tackle the ill-posed 2D-to-3D unprojection inherent in GS-based approaches. Our core insight is that high-fidelity geometric cues can serve as strong guidance to recover the degraded geometric structure during training, thereby mitigating the 3D geometric inconsistencies as established in~\cref{sec:intro}. Inspired by recent hybrid 3DGS designs~\cite{ufo,instantsplat}---combining per-scene optimization for geometric fidelity with feed-forward mechanisms for efficiency and scalability---we leverage per-scene optimization as the source of this precise 3D supervision. To this end, we first revisit the 3DGS optimization process.

\subsubsection{3DGS.}
3DGS represents scenes as a collection of 3D Gaussian primitives, each defined by a mean $\mu$ and covariance $\Sigma$ governing its spatial extent: $G(x) = \exp(-\frac{1}{2}(x-\mu)^\top\Sigma^{-1}(x-\mu))$. To optimize the attributes of these primitives, it employs tile-based rasterization to compute the rendered color $C(v)$ at pixel $v$:
\begin{equation}
C(v) = \sum_{k \in \mathcal{N}} c_k \alpha_k \prod_{j=1}^{k-1} (1-\alpha_j),
\label{eq:3dgs:color}
\end{equation}
where $\mathcal{N}$ denotes the set of Gaussians within the tile, $c_k$ denotes the color of the $k$-th Gaussian and $\alpha_k = o_k G^{2D}_k(v)$ represents the contribution of the $k$-th Gaussian, with $o_k$ being its opacity and $G^{2D}_k(\cdot)$ its 2D projection. The rendered color is subsequently supervised by a photometric loss against the ground truth.

While standard 3DGS yields robust 3D reconstructions that often act as a LiDAR surrogate~\cite{3dgsdet,drivinggaussian,lidargs}, the resulting explicit 3D Gaussians fundamentally misalign with the BEV representation in real-world AD. As shown in~\cref{fig:1_limit}(a,e), vanilla 3DGS faithfully reconstructs near-field camera views (\textit{image-consistency}) driven by the photometric loss; however, its BEV renderings often scatter splats across irrelevant, non-object regions (\textit{BEV-inconsistency})—a critical flaw for reliable BEV perception. This failure is particularly pronounced in driving scenarios characterized by sparse camera coverage and minimal multi-view overlap.

\subsubsection{BAGS.}
Beyond image-level cues, we contend that BEV structural layouts---driven by our 3D projection-first principle---can impose stringent $xy$-plane spatial priors to mitigate the depth ambiguity inherent in sparse 2D supervision. By providing such robust geometric constraints, this novel paradigm offers reliable guidance for precise 3D localization where visual data alone is insufficient. To this end, we propose BAGS, which enforces cross-modal consistency by synchronizing 2D and BEV representations from the correspondence set $\tilde{\mathcal{S}}$. As illustrated in~\cref{fig:2_main}, BAGS prioritizes the learning of object-centric 3D geometry through occupancy maps as opposed to photometric attributes, effectively leveraging the color-agnostic nature of the BEV domain. Specifically, we optimize the accumulated occupancy $O(v)$ via a differentiable rasterization pipeline:
\begin{equation}
O(v) = 1 - \prod_{k \in \mathcal{N}} (1 - \alpha_k).
\label{eq:3dgs:occ}
\end{equation}
The resulting 2D and BEV occupancy maps, $O^{\mathrm{img}}$ and $O^{\mathrm{bev}}$, are aligned with the mask priors in $\tilde{\mathcal{S}}$ via a smooth L1 loss. This joint optimization over instance masks $i \in I$ and BEV instances $b \in B$ ensures cross-modal consistency:
\begin{equation}
\mathcal{L}_{\text{occ}} = \left\| O^{\mathrm{img}}(v) - i(v) \right\|_{\text{SL}_1} + \left\| O^{\mathrm{bev}}(v) - b(v) \right\|_{\text{SL}_1}.
\end{equation}
Moreover, motivated by depth regularization techniques for outdoor scenes~\cite{depthreg,drivinggaussian}, we optimize the per-pixel depth $D(v)$ via the rasterization pipeline:
\begin{equation}
D(v) = \sum_{k \in \mathcal{N}} d_k \alpha_k \prod_{j=1}^{k-1} (1 - \alpha_j),
\label{eq:3dgs:depth}
\end{equation}
where $d_k$ is the camera-space depth of the $k$-th Gaussian center. We supervise $D(v)$ using a dense depth prior $D^*_{\text{dense}}$, obtained by regularizing the predictions of a depth estimator $F_{\theta}$~\cite{zoedepth} with metrically consistent and sparse LiDAR data:
\begin{equation}
\mathcal{L}_{\text{depth}} = \left\| D(v) - D^*_{\text{dense}}(v) \right\|_{1}, \quad \text{where} \;\; D^*_{\text{dense}} = s^* \cdot F_{\theta}(\mathbf{I}) + t^*,
\label{eq:3dgs:loss:depth}
\end{equation}
where $s^*$ and $t^*$ are optimized scale and translation parameters, and $\mathbf{I}$ is the input image. Finally, the total objective for BAGS is defined as:
\begin{align}
\mathcal{L}_{\text{BAGS}} &= (1-\lambda_{\text{ssim}})\mathcal{L}_{\text{color}} + \lambda_{\text{ssim}}\mathcal{L}_{\text{D-SSIM}} + \lambda_{\text{depth}}\mathcal{L}_{\text{depth}} + \lambda_{\text{occ}}\mathcal{L}_{\text{occ}},
\label{eq:3dgs:total_loss}
\end{align}
where $\mathcal{L}_{\text{color}}$ and $\mathcal{L}_{\text{D-SSIM}}$ follow the original 3DGS formulation~\cite{3dgs}. This joint supervision effectively constrains the 3D primitives within a well-defined spatial volume, preventing the geometric collapse often observed in sparse-view settings. To suppress background noise, we mask the image with the instance map $I$, thereby preventing the proliferation of redundant floater Gaussians in irrelevant regions. As shown in~\cref{fig:1_limit}(b,f) and~\ref{fig:2_main}, BAGS yields compact and geometrically faithful 3D Gaussians, establishing a robust source of geometric supervision.

\begin{figure}[tb]
  \begin{minipage}[t]{0.5\linewidth}
    \centering
    \includegraphics[width=\linewidth]{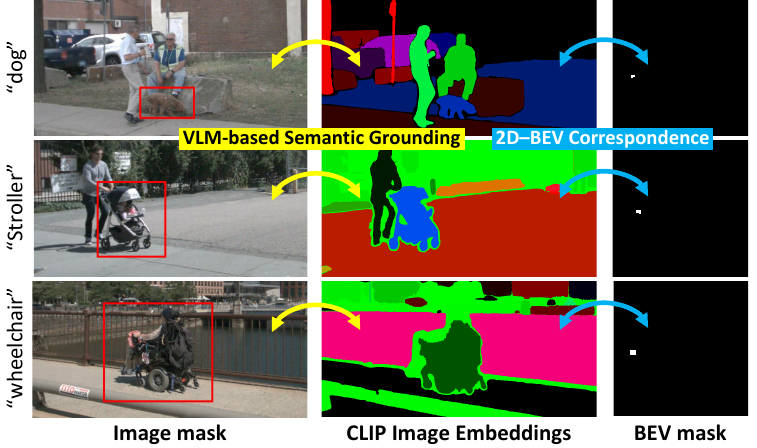}
    \caption{Open-world 2D--BEV auto-labeling.}
    \label{fig:4_oval}
  \end{minipage}
  \hfill
  \begin{minipage}[t]{0.5\linewidth}
    \centering
    \includegraphics[width=\linewidth]{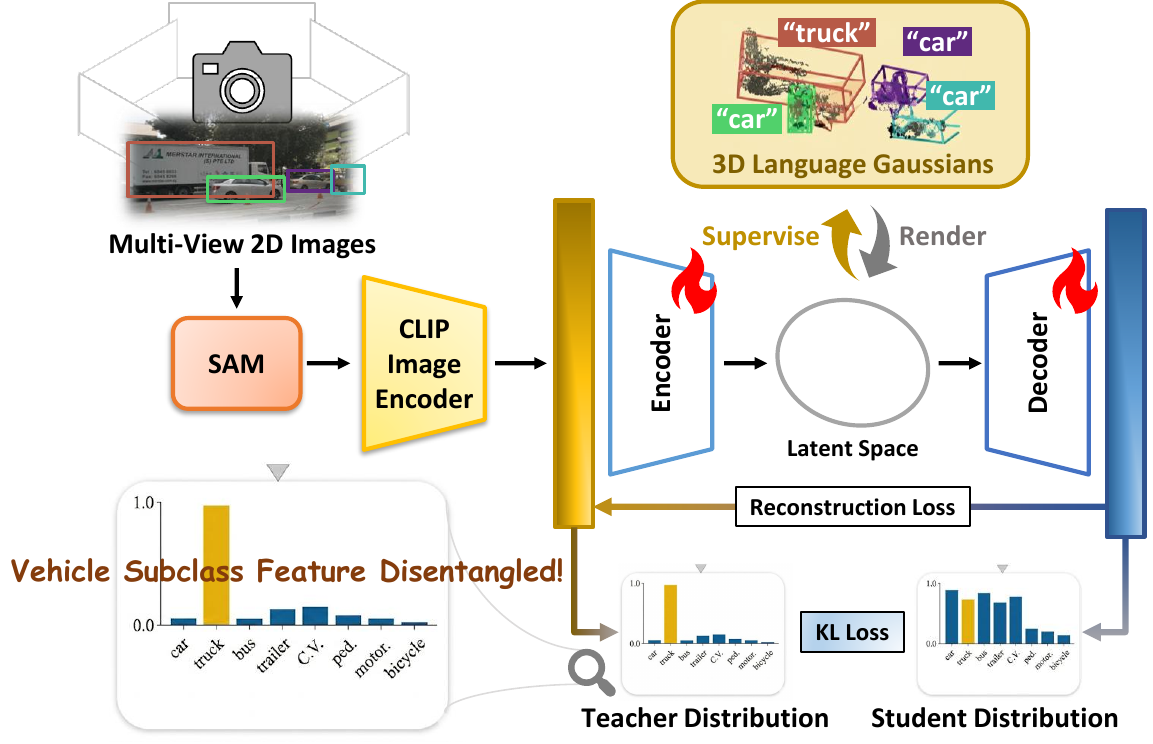}
    \caption{Language-embedded BAGS.}
    \label{fig:4_lang}
  \end{minipage}
\end{figure}

\subsubsection{Language-Embedded BAGS.}
The rising demand for OV 3D scene understanding in AD underscores the extensibility of BAGS, where 3D Gaussians inherit VLM attributes. A key challenge in this task is reducing the memory overhead incurred by processing high-dimensional VLM embeddings during 3DGS optimization. As shown in~\cref{fig:4_lang}, recent work~\cite{langsplat} mitigates this by establishing a lower-dimensional latent space via an encoder-decoder pretraining; it optimizes compact latent features as ground-truth targets. However, this innovative design consistently fails in AD scenarios, where features of vehicle subclasses (\eg, ``car'' and ``truck'') become entangled during the projection into the latent space. Consequently, the decoded features reveal previously unobserved disentanglement problems among semantically similar subclasses.

\begin{figure}[tb]
  \centering
   \includegraphics[width=\linewidth]{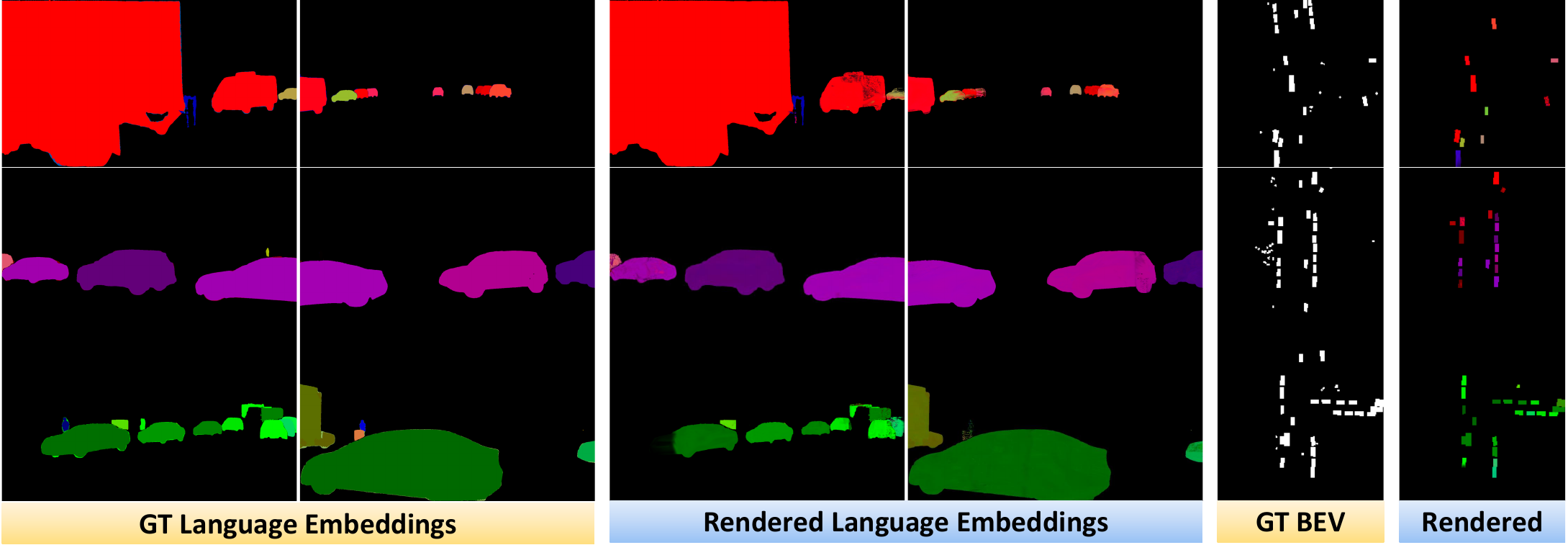}
   \caption{Language-embedded OV 3D scene understanding via the Lang--BAGS protocol.}
   \label{fig:5_langbags}
\end{figure}

To validate the extensibility of language-embedded BAGS---Lang--BAGS---we propose a disentanglement distillation loss that aligns the categorical similarity distributions of reconstructed features with those of original VLM embeddings over subclass embeddings $\mathbf{W}$, inducing discriminative latent projections:
\begin{equation}
\mathcal{L}_{\text{Lang--BAGS}} = \tau^{2} \cdot \text{KL} \left[ \sigma \left( \frac{s \cdot \mathbf{x}_{\text{orig}} \mathbf{W}^\top}{\tau} \right) \parallel \sigma \left( \frac{s \cdot \mathbf{x}_{\text{rec}} \mathbf{W}^\top}{\tau} \right) \right],
\end{equation}
where $\sigma(\cdot)$ denotes the softmax function; $s$ and $\tau$ are the logit scale and temperature, respectively; $\mathbf{x}_{\text{orig}}$ is the original VLM embedding; and $\mathbf{x}_{\text{rec}}$ is the reconstructed feature. As shown in~\cref{fig:4_oval,fig:5_langbags}, Lang--BAGS facilitates the extension of BAGS to OV 3D scene grounding, where its sparse-view robustness maintains language consistency across image and BEV domains.

\subsection{BEV-Aware Gaussian Distillation (BAGD)}
\label{sec:method:c3}

In this section, we introduce OVBEVSeg, a hybrid 3DGS-based framework addressing three key objectives: open-world recognition through PBL, 3D geometric reconstruction via BAGS, and efficient knowledge transfer using BAGD. Following recent hybrid paradigms~\cite{ufo,instantsplat}, we first optimize teacher Gaussians offline with BAGS on temporally downsampled sequences to capture discriminative scene semantics. These high-fidelity primitives provide a rigorous geometric foundation, serving as a dense supervisory signal to guide the student model.

A fundamental challenge in this geometric distillation lies in the inherent lack of one-to-one correspondence between student and teacher Gaussians. While the student predicts a grid-aligned feature tensor $\hat{\mathbf{G}} \in \mathbb{R}^{N \times H_F \times W_F \times d}$, the teacher conversely optimizes an unstructured, dense set of primitives. To bridge this structural gap, we leverage per-pixel opacity $\alpha$ to identify the most influential teacher Gaussian for each ray—a natural choice since objects in driving scenes are rarely stacked vertically in BEV. This allows a single dominant Gaussian to effectively represent the local 3D geometry. Specifically, for each pixel $v$, we select a representative Gaussian $G^*(v) := G_{k^*(v)}$ by maximizing its rendering contribution: $k^*(v) = \arg\max_{k \in \mathcal{N}(v)} \alpha_k$. By mapping the attributes of these selected primitives into the supervision tensor $\mathbf{G}$, we ensure pixel-perfect spatial alignment, effectively anchoring discrete grids to continuous 3D Gaussians.

On the student side, the predicted 3D Gaussians are projected onto the same BEV plane to obtain $\hat{\mathbf{G}}$. We then train the student via a symmetrized distillation objective~\cite{kl} that enforces teacher--student distribution alignment:
\begin{equation}
\mathcal{L}_{\text{BAGD}} = \frac{1}{2|\mathcal{V}|} \sum_{v \in \mathcal{V}} \left[ \text{KL}(\hat{G}(v) \| G^*(v)) + \text{KL}(G^*(v) \| \hat{G}(v)) \right],
\label{eq:dist:loss}
\end{equation}
where $\mathcal{V}$ denotes the set of pixel coordinates. This distillation strategy effectively harmonizes the two 3DGS variants, synergizing their respective strengths to satisfy stringent real-time requirements within a unified OVBS framework.


\section{Experiments}
\label{sec:experiments}

\paragraph{Datasets and evaluation metrics.}
OVBEVSeg is evaluated on the official nuScenes dataset~\cite{nuscenes}, a large-scale AD benchmark with synchronized multi-sensor data across diverse weather and times of day. It contains 1,000 scenes split into 700 training, 150 validation, and 150 test scenes. Each 20-second scene includes data from six cameras providing a full 360-degree view around the ego vehicle. For OVBS, we follow the category split in prior work~\cite{findnpropagate}, dividing vehicle categories into five base and three novel classes. For fair comparison with prior work~\cite{gaussianlss}, we apply a visibility filter that selects objects with at least 40\% coverage. This focuses the evaluation on adequately visible objects. The BEV grid has size (200$\times$200), covering ([-50 m, 50 m]) along both the X (forward) and Y (sideways) axes relative to the ego vehicle. Each cell corresponds to (0.5 m$\times$0.5 m).

\paragraph{Implementation details.}
The online framework operates exclusively on camera inputs with LiDAR priors confined to offline, and is optimized using a weighted sum of a segmentation loss on cosine-similarity logits~\cite{clip} and a KL divergence loss~\cite{kl}, with weights set to 1.0 and 0.01, respectively. For category prompting, we utilize the hand-crafted templates from ViLD~\cite{vild}. Using AdamW~\cite{adamw} with a learning rate of $3 \times 10^{-4}$ and weight decay of $1 \times 10^{-7}$, the model is trained for 50 epochs with a batch size of 8 on two A100 GPUs. Input images are resized to $224 \times 480$ without augmentation, processing multi-scale BEV features at resolutions of $50 \times 50$, $100 \times 100$, and $200 \times 200$. By default, we use an EfficientNet-B4 backbone~\cite{efficientnet} and disable visibility filtering during inference.

\begin{table}[tb]
  \centering
  \caption{Comparison of OVBS in the multi-class setting on the nuScenes \textit{validation} set~\cite{nuscenes}. Note that $\dagger$ denotes visibility filtering. $\Delta_B$ and $\Delta_S$ denote the relative improvements over the baseline~\cite{gaussianlss} and the SOTA model~\cite{tade}, respectively.}
  \label{tab:main_mc}
  \resizebox{\textwidth}{!}{%
  \begin{tabular}{@{}l | cc | c ccc ccccc | c@{}}
    \toprule
    \multirow{2}{*}{\textbf{Method}} &
    \textbf{Memory} &
    \multirow{2}{*}{\textbf{FPS $\uparrow$}} &
    &
    \multicolumn{3}{c}{\textbf{Novel (IoU) $\uparrow$}} &
      \multicolumn{5}{c}{\textbf{Base (IoU) $\uparrow$}} &
      \textbf{Mean} \\
    \cmidrule(lr){5-7}\cmidrule(lr){8-12}
    & \textbf{(MiB) $\downarrow$} & & &
    truck & bus & motorcycle &
    car & trailer & construction\_vehicle & pedestrian & bicycle &
    \textbf{IoU $\uparrow$} \\
    \midrule
    \textit{Fully Supervised}$^{\dagger}$~\cite{gaussianlss} &
    33.0 & 80.2 & &
    29.7 & 38.4 & 10.4 &
    - & - & - & - & - &
    - \\
    \hdashline[1.0pt/2pt]
    \multicolumn{13}{@{}l}{\textit{OVBS}}\\
    \quad VED$^{\dagger}$~\cite{ved} &
    - & - & &
    0.0  & 0.0 & 0.0 &
    7.4  & 0.0 & 0.0 & 0.0 & 0.0 &
    0.9 \\
    \quad VPN$^{\dagger}$~\cite{vpn} &
    - & - & &
    0.0 & 0.0 & 0.0 &
    16.6 & 4.9 & 7.1 & 0.0 & 4.4 &
    4.1 \\
    \quad PON$^{\dagger}$~\cite{pon} &
    38.6 & 43.8 & &
    0.0 & 0.0 & 0.0 &
    24.7 & 16.6 & \underline{12.3} & 8.2  & 9.4  &
    8.9 \\
    \quad DiffBEV$^{\dagger}$~\cite{diffbev} &
    - & - & &
    0.0 & 0.0 & 0.0 &
    38.9 & 21.1 & 8.4  & 9.6  & \underline{13.2} &
    11.4 \\
    \quad GaussianLSS$^{\dagger}$~\cite{gaussianlss} &
    33.0 & 80.2 & &
    0.0 & 0.0 & 0.0 &
    40.0 & 25.1 & 11.6 & \underline{14.4} & 10.4 &
    12.7 \\
    \quad TaDe$^{\dagger}$~\cite{tade} &
    41.5 & 51.4 & &
    0.0 & 0.0 & 0.0 &
    \textbf{42.8} & \underline{26.3} & 11.4 & 14.0 & \textbf{14.2} &
    \underline{13.6} \\
    \quad OVBEVSeg (Ours)$^{\dagger}$ &
    32.4 & 79.6 & &
    \textbf{19.0} & \textbf{20.3} & \textbf{6.6} &
    \underline{41.8} & \textbf{26.9} & \textbf{13.1} & \textbf{15.0} & 12.0 &
    \textbf{19.3} \\
    \quad\quad & & & $\Delta_B$ &
    \scriptsize\textcolor{ForestGreen}{(+19.0)} &
    \scriptsize\textcolor{ForestGreen}{(+20.3)} &
    \scriptsize\textcolor{ForestGreen}{(+6.6)} &
    \scriptsize\textcolor{ForestGreen}{(+1.8)} &
    \scriptsize\textcolor{ForestGreen}{(+1.8)} &
    \scriptsize\textcolor{ForestGreen}{(+1.5)} &
    \scriptsize\textcolor{ForestGreen}{(+0.6)} &
    \scriptsize\textcolor{ForestGreen}{(+1.6)} &
    \scriptsize\textcolor{ForestGreen}{(+6.6)} \\
    \quad\quad & & & $\Delta_S$ &
    \scriptsize\textcolor{ForestGreen}{(+19.0)} &
    \scriptsize\textcolor{ForestGreen}{(+20.3)} &
    \scriptsize\textcolor{ForestGreen}{(+6.6)} &
    \scriptsize\textcolor{BrickRed}{(-1.0)} &
    \scriptsize\textcolor{ForestGreen}{(+0.6)} &
    \scriptsize\textcolor{ForestGreen}{(+1.7)} &
    \scriptsize\textcolor{ForestGreen}{(+1.0)} &
    \scriptsize\textcolor{BrickRed}{(-2.2)} &
    \scriptsize\textcolor{ForestGreen}{(+5.7)} \\
    \bottomrule
  \end{tabular}}
\end{table}

\begin{table}[!t]
    \begin{minipage}[t]{0.487\textwidth}
        \caption{Single-class OVBS comparison. $*$: novel classes withheld during training.}
        \label{tab:main_sc}
        \resizebox{\linewidth}{!}{%
            \begin{tabular}{@{}l | cc | c ccc@{}}
                \toprule
                \multirow{2}{*}{\textbf{Method}} &
                \textbf{Memory} &
                \multirow{2}{*}{\textbf{FPS $\uparrow$}} &
                &
                \multicolumn{3}{c}{\textbf{IoU $\uparrow$}} \\
                \cmidrule(lr){5-7}
                & \textbf{(MiB) $\downarrow$} & & &
                \textbf{vehicle$^*$} &
                \textbf{pedestrian$^{\dagger}$} &
                \textbf{mIoU} \\
                \midrule
                \textit{Fully Supervised}~\cite{gaussianlss} & 33.0 & 80.2 & & 38.3 & 18.0 & 28.1 \\
                \hdashline[1.0pt/2pt]
                \multicolumn{7}{@{}l}{\textit{OVBS}}\\
                \quad CVT~\cite{cvt}             & 35.0  & 107.6 & & 24.6 & 14.2 & 19.4 \\
                \quad FIERY static~\cite{fiery}  & 40.0  & 27.3  & & 28.0 & 17.2 & 22.6 \\
                \quad SimpleBEV~\cite{simplebev} & 331.0 & 37.1  & & 28.9 & 17.1 & 23.3 \\
                \quad GaussianLSS~\cite{gaussianlss} & 33.0 & 80.2 & & 29.7 & 18.0 & 23.9 \\
                \quad PointBeV~\cite{pointbev}   & 126.0 & 32.0  & & \underline{30.3} & \textbf{18.5} & \underline{24.4} \\
                \quad OVBEVSeg (Ours)            & 28.7  & 80.2  & & \textbf{34.7} & \underline{18.4} & \textbf{26.6} \\
                \quad\quad & & & $\Delta_B$ &
                    \scriptsize\textcolor{ForestGreen}{(+5.0)} &
                    \scriptsize\textcolor{ForestGreen}{(+0.4)} &
                    \scriptsize\textcolor{ForestGreen}{(+2.7)} \\
                \quad\quad & & & $\Delta_S$ &
                    \scriptsize\textcolor{ForestGreen}{(+4.4)} &
                    \scriptsize\textcolor{BrickRed}{(-0.1)} &
                    \scriptsize\textcolor{ForestGreen}{(+2.2)} \\
                \bottomrule
            \end{tabular}%
        }
    \end{minipage}%
    \hfill
    \begin{minipage}[t]{0.473\textwidth}
        \caption{OVBS results with self- and semi-supervised baselines on nuScenes.}
        \label{tab:semi}
        \resizebox{\linewidth}{!}{%
            \begin{tabular}{@{}clcccc@{}}
                \toprule
                \textbf{GT Ratio} & \textbf{Method} & truck & bus & motorcycle & \textbf{mIoU} $\uparrow$ \\
                \midrule
                \multirow{2}{*}{0\%}
                  & 2D-OV-then-BEV    & 3.7 & 2.6 & 0.0 & 2.1 \\
                  & \cellcolor{yellow!20}\textbf{OVBEVSeg (Ours)}
                    & \cellcolor{yellow!20}19.0
                    & \cellcolor{yellow!20}\underline{20.3}
                    & \cellcolor{yellow!20}6.6 
                    & \cellcolor{yellow!20}\underline{15.3} \\
                \midrule
                \multirow{3}{*}{$(1, 5\%]$}
                  & LetsMap~\cite{self}     & 13.6 & - & 5.8 & 9.7 \\
                  & UniMatch~\cite{unimatch} & 10.4 & 5.7 & 1.6 & 5.9 \\
                  & Semi-BEVseg~\cite{semi} & 14.7 & 9.8 & 1.6 & 8.7 \\
                \midrule
                \multirow{2}{*}{20\%}
                  & UniMatch~\cite{unimatch} & 16.8 & 10.5 & 2.6 & 10.0 \\
                  & Semi-BEVseg~\cite{semi} & \underline{20.4} & 16.8 & \underline{6.7} & 14.6 \\
                \midrule
                \multirow{2}{*}{40\%}
                  & UniMatch~\cite{unimatch} & 18.0 & 17.0 & 5.6 & 13.5 \\
                  & Semi-BEVseg~\cite{semi} & \textbf{22.7} & \textbf{21.6} & \textbf{6.8} & \textbf{17.0} \\
                \bottomrule
            \end{tabular}%
        }
    \end{minipage}
\end{table}

\subsection{Quantitative Research}

\paragraph{Comparison with closed-set methods.}
We evaluate OVBEVSeg against state-of-the-art (SOTA) closed-set BEV segmentation models on nuScenes~\cite{nuscenes} under OVBS settings, focusing on novel classes. In~\cref{tab:main_mc,tab:main_sc}, SC and MC denote training on a single class (\eg, ``vehicle'') and multiple classes jointly, respectively; notably, the ``vehicle'' category in SC uses only base-class annotations. Leveraging our BEV pseudo-labels, OVBEVSeg sets a new SOTA on novel classes, surpassing previous projection-based methods~\cite{pointbev,tade} by 4.4 and 15.3 mIoU in SC and MC, while being 2.7$\times$ faster and consuming only 0.22$\times$ the memory. It also outperforms the feed-forward 3DGS-based baseline~\cite{gaussianlss} by 2.7 and 6.6 mIoU, while maintaining competitive online efficiency. These results strongly validate that our OVBS framework paves the way for practical deployment.

\paragraph{Comparison with semi- and self-supervised methods.}
Despite the clear necessity of OVBS, the current scarcity of direct competitors in this emerging domain leads us to broaden our evaluation. We include two additional baselines with objectives relevant to OVBS. First, we compare against self- and semi-supervised methods~\cite{self,unimatch,semi} trained with varying ground-truth (GT) ratios. In~\cref{tab:semi}, our method—operating with 0\% GT—significantly outperforms these baselines with $<20\%$ GT, and remains competitive with those using up to 40\%. Second, a ``2D-OV-then-BEV'' pipeline—comprising a LiDAR-regularized depth estimator~\cite{depthreg} and a strong OV detector~\cite{groundingdino}—is evaluated as a naive OVBS baseline. Our approach markedly outperforms this alternative, particularly in sparse-view settings where 2D unprojection errors are amplified. These fairness comparisons underscore the need for our method in the critical yet underserved OVBS task.


\begin{table}[tb]
  \centering
  \begin{minipage}[tb]{0.38\textwidth}
      \centering
      \caption{OVBS comparison with different 3D priors and non-vehicle classes (4 novel classes).}
      \label{tab:rebut}
      \resizebox{\linewidth}{!}{
      \begin{tabular}{@{}lcccc@{}}
        \toprule
        \multirow{2}{*}{\textbf{Method}} & \multicolumn{4}{c}{\textbf{Novel (IoU) $\uparrow$}} \\
        \cmidrule(lr){2-5}
        & truck & bus & motorcycle & cone \\
        \midrule
        UNION~\cite{union} & \textbf{19.0} & \textbf{20.3} & 6.6 & - \\
        F\&B~\cite{findnpropagate} & 14.2 & 15.5 & \textbf{7.9} & \textbf{11.5} \\
        \bottomrule
      \end{tabular}}
\end{minipage}
  \hfill
  \begin{minipage}[tb]{0.6\textwidth}
    \centering
    \caption{Per-sample submodule runtime on an A100 GPU. \colorbox{cyan!10}{Inference-only} modules are shared with the baseline. $\natural$ uses 10\% of per-scene samples.}
    \label{tab:time}
    \resizebox{\linewidth}{!}{%
    \begin{tabular}{@{}lccccc@{}}
      \toprule
      \textbf{Online Method} &
      \cellcolor{cyan!10}\textbf{Backbone} &
      \cellcolor{cyan!10}\textbf{Neck} &
      \cellcolor{cyan!10}\textbf{Encoder} &
      \cellcolor{cyan!10}\textbf{Head} &
      \textbf{BAGD}$^\natural$ (ms) \\
      \midrule
      Ours (BAGD$^\natural$) &
      \cellcolor{cyan!10}29.3 &
      \cellcolor{cyan!10}5.6 &
      \cellcolor{cyan!10}5.7 &
      \cellcolor{cyan!10}1.38 &
      0.37 \\
      \bottomrule
      \toprule
      \textbf{Offline Method} & \textbf{3D DET} & \textbf{SAM} & \textbf{CLIP} & \textbf{BAGS}$^\natural$ & \textbf{Total} (min) \\
      \midrule
      Ours (PBL + BAGS$^\natural$) & $<$ 0.01 & 0.23 & 0.32 & 1.54 & 2.1 \\
      \bottomrule
    \end{tabular}}
  \end{minipage}
\end{table}

\subsection{Ablation Study}

\paragraph{Ablation of 3D priors and non-vehicle classes.}
\Cref{tab:rebut} shows that PBL consistently improves OVBS across unsupervised~\cite{union} and VLM-based OV priors~\cite{findnpropagate}, while reflecting their complementary strengths, such as better recognition of small static objects with the OV prior. This highlights the generality of PBL beyond vehicle-centric categories and its compatibility with future 3D priors.

\paragraph{Submodule runtime.}
\Cref{tab:time} details the per-scene submodule runtimes on an A100 GPU. Our model extends the baseline with a lightweight BAGD module completely removed during inference. The remaining components (\eg, PBL, BAGS, and cached 3D detections) are precomputed and cached offline prior to online training. Each submodule incurs negligible computational overhead via asynchronous batch parallelism---SAM producing masks for batch-wise CLIP aggregation---with BAGS leveraging our online efficient distillation scheme. The offline preprocessing of the entire training set—roughly 30,000 samples—takes approximately 4 days using 8 A100 GPUs, while online efficiency remains on par with the baseline. These results highlight that OVBEVSeg significantly boosts OVBS performance while preserving competitive computational online efficiency.

\paragraph{Ablation of the individual modules.}
\Cref{tab:ablation} evaluates the contribution of each component within OVBEVSeg. While Group II establishes a new SOTA for novel classes, thereby facilitating OV generalization, the substantial performance degradation in Group III underscores a critical limitation: integrating vanilla 3DGS supervision with BAGD propagates inaccurate 3D geometry, leaving the 2D unprojection dilemma unresolved. In contrast, Group IV demonstrates that combining BAGS with BAGD ensures geometric consistency across camera and BEV views, providing constructive supervision that boosts overall OVBS performance by refining boundary details across all classes. These results confirm that the synergistic integration of all proposed modules yields a robust OVBS framework for real-world driving scenarios.

\begin{table}[tb]
  \centering
  \begin{minipage}[tb]{0.36\textwidth}
    \centering
    \caption{Component ablation.}
    \label{tab:ablation}
    \resizebox{\linewidth}{!}{%
    \begin{tabular}{@{}ccccccc@{}}
      \toprule
      \multirow{2}{*}{\textbf{Group}} &
      \multicolumn{3}{c}{\textbf{Components}} &
      \multicolumn{3}{c@{}}{\textbf{IoU $\uparrow$}} \\
      \cmidrule(lr){2-4}\cmidrule(lr){5-7} 
       & PBL & BAGS & BAGD & Novel & Base & Mean \\
      \midrule
      I   &                      &      &      & 0.0 & 20.3 & 10.2 \\
      II & \checkmark            &      &      & 14.0 & 20.5 & 17.3 \\
      III  & \checkmark          &      & \checkmark      & 6.1 & 9.8 & 7.9 \\
      IV   & \checkmark          & \checkmark & \checkmark & \textbf{15.3} & \textbf{21.8} & \textbf{19.3} \\
      \bottomrule
    \end{tabular}}
  \end{minipage}
  \hfill
  \begin{minipage}[tb]{0.28\textwidth}
    \centering
    \caption{BAGD ratio.}
    \label{tab:bagd}
    \resizebox{0.8\linewidth}{!}{
    \begin{tabular}{@{}cccc@{}}
      \toprule
      \multirow{2}{*}{Ratio} & \multicolumn{3}{c}{\textbf{IoU $\uparrow$}} \\
      \cmidrule(lr){2-4}
                                          & Novel & Base & Mean \\
      \midrule
      0    & 14.0 & 20.5 & 17.8 \\
      0.05 & 14.8 & 21.5 & 18.9 \\
      0.1  & \textbf{15.3} & \textbf{21.8} & \textbf{19.3} \\
      \bottomrule
    \end{tabular}}
  \end{minipage}
  \hfill
  \begin{minipage}[tb]{0.34\textwidth}
    \centering
    \caption{Map segmentation.}
    \label{tab:map}
    \resizebox{\linewidth}{!}{
    \begin{tabular}{@{}lccc@{}}
      \toprule
      \textbf{Method} & drivable & crossing & walkway \\
      \midrule
      TaDe        & 65.9 & 40.9 & 42.3 \\
      CVT         & 74.3 & 36.8 & 39.9 \\
      LSS         & 75.4 & 38.8 & 46.3 \\
      GaussianLSS & \underline{76.3} & \underline{46.3} & \underline{50.2} \\
      \cellcolor{yellow!20}\textbf{Ours}
        & \cellcolor{yellow!20}\textbf{78.1}
        & \cellcolor{yellow!20}\textbf{49.4}
        & \cellcolor{yellow!20}\textbf{51.9} \\
      \bottomrule
    \end{tabular}}
  \end{minipage}
\end{table}
\begin{figure}[tb]
  \centering
   \includegraphics[width=\linewidth]{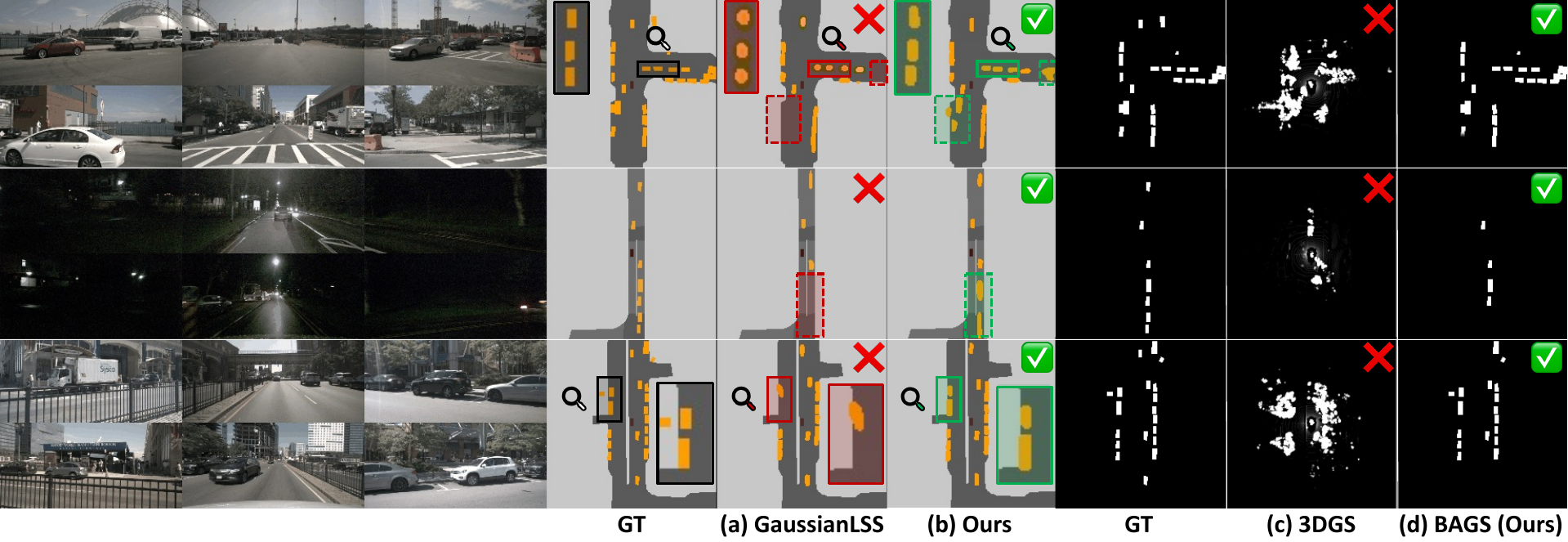}
   \caption{\textbf{(a–b)} Our BEV predictions vs. the baseline: orange regions denote both base and novel vehicle classes. \textbf{(c–d)} Our BEV renderings (BAGS) vs. vanilla 3DGS.}
   \label{fig:3_qual}
\end{figure}

\paragraph{Distillation ratio.}
We evaluate various BAGD ratios to assess their efficacy in 3D geometric knowledge transfer. A typical nuScenes~\cite{nuscenes} scene consists of approximately 40 frames captured at 2 Hz. Given the high temporal redundancy between consecutive frames~\cite{4dgs,bdd100k}, we sample frames at uniform intervals to mitigate 3DGS optimization overhead. This selective distillation strategy enhances the online efficiency of BAGS by focusing on key scene-discriminative frames. As shown in~\cref{tab:bagd}, OVBS performance scales near-linearly with the distillation ratio. Notably, a ratio of 0.1 (\ie, 4 frames per scene) achieves the optimal efficiency--accuracy balance, underscoring our framework's robust scalability.

\paragraph{Map segmentation.}
\Cref{tab:map} evaluates map classes requiring precise road geometry. Comparing our BAGD-enhanced model against existing methods, we predict categories such as drivable areas, pedestrian crossings, and walkways following the same setup of prior work~\cite{gaussianlss}. Our approach achieves SOTA performance across all classes, effectively reinforcing the underlying scene structure.

\subsection{Qualitative Research}

\paragraph{Qualitative results.}
As illustrated in~\cref{fig:3_qual}, the baseline~\cite{gaussianlss} \textbf{(a)} fails to detect novel-class objects (\eg, ``truck'') and collapses geometries into unstructured, amorphous volumes, compromising reliability in safety-critical AD. In contrast, OVBEVSeg \textbf{(b)} successfully captures novel classes with sharp boundaries, underscoring its efficacy for OVBS. Regarding geometric fidelity under sparse ego-car viewpoints, vanilla 3DGS \textbf{(c)} suffers from inaccurate geometry and misaligned BEV renderings, whereas BAGS \textbf{(d)} effectively resolves these inconsistencies, ensuring robustly aligned 3D geometries as reliable OVBS supervision.

\paragraph{Open-world 2D--BEV auto-labeling.}
Given the increasing scale of datasets, automatic multimodal annotation is essential for real-world AI applications like AD. In contrast to closed-set approaches, our PBL system facilitates OV auto-labeling (\eg, ``stroller'' in~\cref{fig:4_oval}. By projecting cross-modal similarity signals into BEV space via the precomputed correspondence set, our system successfully yields synchronized 3D, 2D, and BEV annotations. This framework establishes a scalable pipeline for high-fidelity, multimodal OV auto-labeling in 3D.

\paragraph{Open-world 3D scene understanding via 3DGS.}
\Cref{fig:3_qual} demonstrates that Lang--BAGS effectively embeds CLIP features into 3D Gaussians, ensuring their projections in both image and BEV domains consistently align with GT embeddings. Beyond standard 3DGS, BAGS enables language-aware 3D scene understanding in AD scenarios, maintaining robustness even with sparse viewpoints.


\section{Conclusion}
In this work, we introduced open-vocabulary BEV segmentation (OVBS) to address the inherent safety-critical limitations of closed-set driving systems. To mitigate the 3D geometric inconsistencies common in 2D-to-3D semantic lifting---identified as a primary obstacle in OVBS---we proposed OVBEVSeg, the first OVBS framework that enhances Gaussian splatting (GS)-based unprojection by leveraging robust 3D geometric constraints through three synergistic modules---(1) PBL: pseudo-BEV labeling via unsupervised 3D projection for OV generalization; (2) BAGS: 2D--BEV-consistent 3D Gaussian splatting for 3D geometry recovery; and (3) BAGD: efficiency-preserving geometric distillation. Extensive evaluations validate that OVBEVSeg achieves state-of-the-art novel-class recognition with a competitive memory footprint and real-time speed. By enforcing geometric consistency across the 2D-3D-BEV continuum, this work establishes a scalable foundation for robust open-world multimodal 3D perception.
\bibliographystyle{splncs04}
\bibliography{main}

\title{Open-Vocabulary BEV Segmentation with 3D-Aware Geometric Constraints\\(Supplementary Material)}

\titlerunning{OVBEVSeg}

\author{Hojun Choi\inst{1}$^*$\orcidlink{0009-0009-4121-4519}\and
Seulbin Hwang\inst{2}\orcidlink{0009-0009-3448-1862} \and
Daejung Kim\inst{2}\orcidlink{0000-0002-5517-1179} \and
Kisung Kim\inst{2}\orcidlink{0009-0006-8729-6458} \and\\
Hyunjung Shim\inst{1}$^\dagger$\orcidlink{0000-0001-6796-1058} \and
Jinhan Lee\inst{2}$^\dagger$\orcidlink{0009-0003-2965-143X}}

\authorrunning{H.~Choi et al.}

\institute{KAIST AI, South Korea\\
\email{\{hchoi256,kateshim\}@kaist.ac.kr} \and
NAVER LABS, South Korea\\
\email{\{h.sb,daejung.kim,ks.kim,jinhan.lee\}@naverlabs.com}}

\maketitle

\renewcommand{\thesection}{\Alph{section}}
\renewcommand{\thetable}{A\arabic{table}}
\renewcommand{\thefigure}{A\arabic{figure}}
\renewcommand{\theequation}{A\arabic{equation}}


\section*{Contents}
\noindent
A. Device Information \dotfill 20\\
B. Limitations \& Future Work \dotfill 21\\
C. Implementation Details \dotfill 22\\
D. Additional Ablation Study \dotfill 23\\
\hspace*{1em}D.1 PBL: Pseudo-Label Statistics \dotfill 23\\
\hspace*{1em}D.2 BAGS: Comparative Analysis of Loss Formulations \dotfill 24\\
\hspace*{1em}D.3 Lang–BAGS: Disentanglement Analysis \dotfill 26\\
\hspace*{1em}D.4 BAGD: Extension to Top-K Gaussian Gating Mechanism \dotfill 26\\
\hspace*{1em}D.5 Open-Vocabulary 3D Object Detection \dotfill 27\\
E. Additional Qualitative Results \dotfill 28\\
F. Specifications of Baseline Models \dotfill 30\\
G. Specifications of View Transformation \dotfill 32\\
\hspace*{1em}G.1 3D Camera Projection \dotfill 32\\
\hspace*{1em}G.2 BEV Splatting: BEV Features via Gaussian Splatting \dotfill 32\\

\section{Device Information}
\label{appendix:device}
All experiments were implemented in PyTorch 2.1.0 and conducted on a workstation equipped with eight NVIDIA A100-SXM4-80GB GPUs. The online training phase was completed in approximately 6 hours on two GPUs, with a peak VRAM utilization of 21 GB per device. To ensure reproducibility and a fair comparison, we aligned our experimental configuration with the baseline~\cite{gaussianlss} by fixing the random seed to 2025 for all trials. Moreover, the offline stage---leveraging our efficient asynchronous design across the entire nuScenes benchmark---required approximately four days to complete using a distributed configuration of eight GPUs. This offline execution utilized the full GPU memory capacity via multi-threaded processing across all 8 GPUs.

\section{Limitations \& Future Work}
\label{appendix:limitations}
In this section, we provide a critical analysis of the current constraints of our proposed framework, OVBEVSeg. While effective, the system encounters three primary challenges: \textbf{(L1)} the inherent computational overhead of expensive per-scene optimization (\cref{limit:sec:3dgs}) and \textbf{(L2)} the absence of spatiotemporal (4D) consistency across sequential frames (\cref{limit:sec:4d}). Addressing these bottlenecks remains a high-priority objective for the community.


\subsection{Tedious Per-Scene Optimization}
\label{limit:sec:3dgs}
Our methodology utilizes a Gaussian splatting (GS)-based unprojection framework for BEV perception, seamlessly integrated with an online feed-forward 3DGS module~\cite{3dgs,gaussianlss} to ensure high-performance, real-time inference throughput. While 3D geometric representations are subsequently enhanced through offline per-scene optimization, this hybrid 3DGS-based architectural choice retains the necessity for dense 3D geometric supervision. Specifically, achieving peak performance requires intensive optimization on a subset of frames (\eg, 10\% per scene) to ensure high-fidelity reconstruction in visually discriminative environments. While the per-sample optimization latency remains constant across both vanilla 3DGS and BAGS, the latter achieves superior computational efficiency by significantly reducing the training set cardinality required for high-fidelity reconstruction, as established in~\cref{tab:time,tab:bagd}.

Nonetheless, the heavy optimization requirements pose a bottleneck for zero-shot generalization to large-scale environments, a common hurdle in the 3DGS paradigm. We envision that integrating efficiency---focused primitives—as seen in RaSGS~\cite{raings}, Scaffold-GS~\cite{scaffoldgs}, and Mini-GS~\cite{minigs}---will drastically improve the practical deployment of BAGS. As our framework is architecture-agnostic, it can be readily adapted to these emerging 3DGS variants to achieve superior scalability without compromising reconstruction fidelity.

\subsection{Extension to 4D Temporal Dynamics}
\label{limit:sec:4d}
Despite its robust spatial reasoning capabilities, the current architectural design of OVBEVSeg does not yet support temporal consistency or 4D scene reconstruction across sequential frames. Dynamic scene understanding and multi-frame consistency are critical for safety-critical autonomous driving (AD) applications~\cite{drivinggaussian,hugs}. Drawing inspiration from recent advancements in 4DGS~\cite{4dgs}, future work will prioritize the integration of our framework with temporal modeling architectures---a natural extension of 3DGS. This synergy will enable the model to leverage our high-fidelity spatial refinements to effectively address dynamic agents and maintain spatiotemporal consistency across long-range temporal dependencies. Such an integration would enable high-fidelity 3D scene reconstruction from sparse viewpoints across both spatial and temporal dimensions, significantly broadening its utility for AD simulation and closed-loop testing.

\section{Implementation Details}
In this section, we delineate the comprehensive implementation details of our proposed framework (\eg, the specific model configurations).

\subsubsection{Online framework.}
Our online architecture is built upon GaussianLSS~\cite{gaussianlss}, which leverages feed-forward 3DGS to treat BEV feature prediction as a differentiable, GPU-accelerated rasterization process. This design significantly enhances inference throughput (FPS) while minimizing memory overhead---a cornerstone of efficient AD. The optimization objective is formulated as a weighted multi-task loss, comprising a segmentation objective applied to cosine-similarity logits~\cite{clip} and a KL divergence distillation loss~\cite{kl}. We set the respective loss weights to $\lambda_{\text{ovseg}} = 1.0$ and $\lambda_{\text{BAGD}} = 0.01$, ensuring a stable balance between categorical discriminability and geometric alignment. The optimization is conducted for 180,000 iterations ($\approx$ 50 epochs) using the AdamW optimizer~\cite{adamw} with an initial learning rate of $3\times10^{-4}$, weight decay of $1\times10^{-7}$, a cosine learning rate scheduler, and a batch size of 8. Model checkpoints are evaluated every 500 iterations, with the final selection based on peak validation performance. Notably, we omit data augmentation to maintain the structural integrity of the spatial distributions, thereby ensuring the consistency of BAGD. Notably, our camera-only online training eliminates the dependency on LiDAR-derived priors; while such data can be optionally incorporated during the offline phase. This distinguishes our framework from existing multi-sensor BEV approaches that rely on sparse LiDAR supervision to achieve marginally higher performance, offering a more flexible and scalable solution for deployment.

Following the standard open-vocabulary (OV) evaluation protocol~\cite{ov}, GT annotations for the unseen semantic categories (\ie, ``truck,'' ``bus,'' and ``motorcycle'') are strictly excluded during the training phase. To enable OV capabilities, we substitute the standard closed-set segmentation head with an OV head. In this configuration, logits are computed as the cosine similarity between the predicted BEV embeddings and text embeddings derived from a pre-trained vision-language model (VLM)~\cite{clip}. Accordingly, we project the BEV feature dimensionality from 128 to 512 to align with the CLIP embedding space. We utilize dataset-specific class names refined with hand-crafted prompts from ViLD~\cite{vild} to ensure robust semantic grounding. During the training phase, the VLM backbone remains frozen, while the BEV encoder and the OV head are optimized.

\subsubsection{Offline Framework}
Our offline framework comprises two primary components: PBL and BAGS. Utilizing parallelized multi-threaded processing across eight GPUs, the aggregate offline time cost of the entire training set---encompassing 700 scenes equivalent to approximately 30,000 temporal samples (roughly equivalent to 180,000 images)---is approximately 4 days, as detailed in~\cref{tab:time}.

\paragraph{PBL.}
PBL integrates three foundational models: Grounding-DINO~\cite{groundingdino} (base variant) and SAM~\cite{sam} with a ViT-base backbone for open-world object localization, CLIP-ViT-B/16 for semantic feature alignment, and an unsupervised 3D detector~\cite{union}. To maximize throughput, we implement an asynchronous producer-consumer architecture on a single GPU, where SAM serves as the producer for object proposals and CLIP functions as the consumer for feature extraction. This cost-effective design allows for the efficient processing of batched object instances (batch size of 256), fully saturating the available GPU VRAM and significantly reducing offline pseudo-labeling latency.

\paragraph{BAGS.}
BAGS is developed upon the 3DGS codebase with integrated depth regularization~\cite{3dgs,langsplat,depthreg}. In~\cref{eq:3dgs:total_loss}, the loss hyperparameters are empirically set to $\lambda_{\text{color}} = 1.0$, $\lambda_{\text{D-SSIM}} = 1.0$, $\lambda_{\text{depth}} = 1.0$, and $\lambda_{\text{occ}} = 1.0$. We have engineered a customized CUDA-based rasterization pipeline to incorporate learnable occupancy maps, facilitating the joint optimization of all Gaussian parameters.

\section{Additional Ablation Study}
In this section, we provide comprehensive ablation studies to further validate the efficacy of each component in our proposed framework.

\begin{table}[tb]
    \centering
    \caption{\textbf{Pseudo-label statistics on the nuScenes train split.} Comparative results between our pseudo-labels and ground-truth annotations are provided for the three novel classes. All values correspond to total instance counts across the dataset.}
    \label{tab:pbl_stat} 
    
    \begin{tabular}{lccc} 
        \toprule
        \textbf{Metric} & \textbf{Pseudo} & \textbf{GT} & \textbf{Ratio} (\%) \\
        \midrule
        \# Sample Scenes & 24,900 & 28,130 & 88.5 \\
        \hdashline[1.0pt/2pt]
        \textit{Novel classes} & & & \\
        \quad truck      & 23,456 & 72,815 & 32.2 \\
        \quad bus        & 10,205 & 13,163 & 77.5 \\
        \quad motorcycle & 4,169  & 10,109 & 41.2 \\
        \midrule
        \textbf{Total}            & 37,830 & 96,087 & 39.4 \\
        \bottomrule
    \end{tabular}
\end{table}

\subsection{PBL: Pseudo-Label Statistics and Categorical Fidelity}
\paragraph{Pseudo-label statistics.}
\Cref{tab:pbl_stat} provides a comparative analysis between the BEV pseudo-labels generated by our proposed PBL pipeline and the corresponding GT annotations. Our method successfully generates pseudo-labels for 88.5\% of the GT scenes, achieving an overall semantic coverage of 39.4\% relative to the GT annotations. These statistics indicate that the PBL pipeline consistently assigns a substantial volume of high-quality pseudo-labels for novel classes across a wide range of diverse driving environments.

\paragraph{Per-category fidelity analysis.}
At the category level, PBL maintains a high annotation density, generating at least 30\% as many pseudo-labels as GT instances for every novel class. Notably, the ``bus'' class achieves a density of 77.5\%, directly correlating with the significant performance gains observed for this category in~\cref{tab:main_mc}. Specifically, \Cref{tab:pseudo} further details the overall pseudo-label fidelity.

\paragraph{Pseudo-label quality.}
\Cref{tab:pseudo} evaluates the quality of our BEV pseudo-labels on the nuScenes training set, achieving 13.5 and 13.0 mIoU on novel classes with and without visibility filtering, respectively. This underscores the efficacy of PBL as a robust source for OVBS, even in the absence of novel-class ground truth.

\begin{table}[t]
  \centering
  \caption{Novel-class pseudo-label quality of our method on the nuScenes~\cite{nuscenes} training set. $\dagger$: visibility filtering.}
  \label{tab:pseudo}
  \resizebox{0.45\textwidth}{!}{
  \begin{tabular}{@{}lcccc@{}}
    \toprule
    \multirow{2}{*}{\textbf{Method}} & \multicolumn{3}{c}{\textbf{Novel (IoU) $\uparrow$}} & \textbf{Mean} \\
    \cmidrule(lr){2-4}
    & truck & bus & motorcycle & \textbf{IoU} $\uparrow$ \\
    \midrule
    Ours & 15.8 & 17.98 & 5.34 & 13.0 \\
    Ours$^{\dagger}$ & 17.1 & 18.35 & 5.09 & 13.5 \\
    \bottomrule
  \end{tabular}}
\end{table}

\subsection{BAGS: Comparative Analysis of Loss Formulations}
We study how loss formulations affect joint BAGS optimization across image and BEV domains. The results show that the loss choice is crucial for balancing image reconstruction quality with accurate BEV spatial grounding.

\subsubsection{L1 loss.}
\cref{fig:sup:loss}-a shows that a standard $L_1$ loss produces fragmented and discontinuous BEV reconstructions, failing to form a cohesive structure even after 30K optimization iterations. Compared to $L_{\mathrm{BCE}}$, $L_1$ exhibits slower and less stable convergence, as its element-wise regression objective provides limited guidance for enforcing global occupancy consistency. As a result, the optimized 3D Gaussians struggle to maintain coherent density in 3D space, leading to spatial gaps, smearing artifacts, and incomplete object interiors in BEV space. These results indicate that simply increasing the iteration count cannot compensate for the limited cross-modal geometric consistency of the $L_1$ objective.

\subsubsection{Binary cross entropy (BCE).}
In~\cref{fig:sup:loss}-b, $L_{\mathrm{BCE}}$ tends to favor the relatively simple BEV-space binary supervision over the more complex image-space reconstruction. Since the BEV target provides direct occupancy signals, the optimization can quickly reduce the loss by prioritizing coarse BEV structural consistency while under-utilizing image-view cues. Consequently, the 3D Gaussians preserve plausible BEV geometry but fail to reconstruct photometric information in the image domain. This premature convergence shows that $L_{\mathrm{BCE}}$ alone is insufficient for balancing robust BEV grounding with image-space alignment.

\subsubsection{Smooth L1 loss.}
\begin{figure}[tb]
  \centering
  \includegraphics[width=\linewidth,height=\textheight,keepaspectratio]{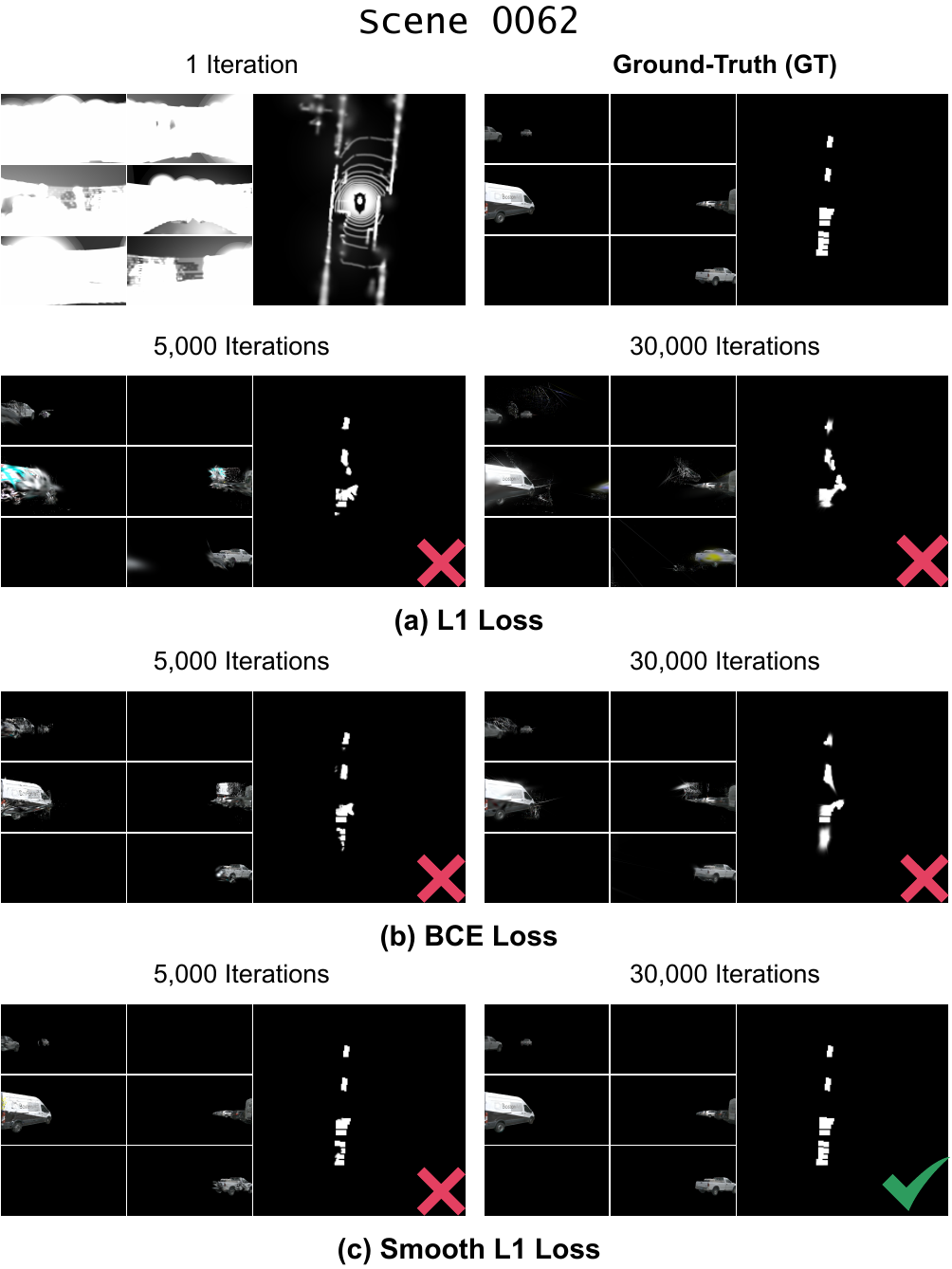}
  \caption{Qualitative validation of 2D--BEV consistency under various losses via BAGS.}
  \label{fig:sup:loss}
\end{figure}

In~\cref{fig:sup:loss}-c, smooth L1 loss provides a balanced objective for joint image and BEV reconstruction. By treating 3D density as continuous regression rather than binary classification, it provides stable gradients while remaining robust to outliers. This better aligns continuous image-space features with discrete BEV occupancy supervision, allowing the optimized 3D Gaussians to preserve cohesive BEV structures and high-fidelity image-view reconstruction.

\clearpage

\begin{table}[tb]
    \centering
    \caption{\textbf{Quantitative analysis of our disentanglement strategy.} Hit Rate represents instances where teacher and student embeddings share identical class labels.}
    \label{tab:sup:disentangle}
    \resizebox{0.45\textwidth}{!}{%
        \begin{tabular}{lc}
            \toprule
            \textbf{Method} & \textbf{Hit Rate} (\%) \\
            \midrule
            LangSplat~\cite{langsplat} & 43.7 \\
            \rowcolor{gold}
            Lang--BAGS (Ours) & 99.4 \\
            \bottomrule
        \end{tabular}
    }
\end{table}

\subsection{Lang--BAGS: Disentanglement Analysis}
This section provides a quantitative validation of the disentanglement strategy of our Lang--BAGS, as established in~\cref{sec:method:c2}. To evaluate the representational fidelity of our framework, we conduct a comparative analysis between the original CLIP feature embeddings (\textit{teacher}) and the reconstructed embeddings (\textit{student}) for each instance mask extracted via our correspondence set across the entire training corpus. The alignment is measured by mapping both embedding sets to the predefined CLIP textual embeddings of the target category set. We quantify the class-wise agreement by identifying ``hits'' where both the teacher and student embeddings yield the identical semantic label. Conversely, a ``miss'' is recorded if the decoded student embedding deviates from the teacher's classification.

As summarized in~\cref{tab:sup:disentangle}, the baseline~\cite{langsplat} is frequently susceptible to semantic entanglement, particularly among vehicular subcategories, resulting in lower alignment scores. In particular, the decoded student embeddings frequently exhibit semantic collapse, gravitating toward the ``car'' category---the most pervasive vehicular representative. This phenomenon is a direct consequence of representational degradation, where the distinct features of minority subclasses, such as ``bus'' or ``truck,'' are subsumed by the dominant class distribution during the feature distillation process. In contrast, our method effectively resolves these ambiguities, achieving a near-perfect hit rate of approximately 100\%. These results empirically demonstrate the efficacy of our disentanglement strategy in preserving high-fidelity semantic information within the geometric primitives.

\subsection{BAGD: Extension to Top-K Gaussian Gating Mechanism}
\begin{figure}[tb]
  \centering
   \includegraphics[width=\linewidth]{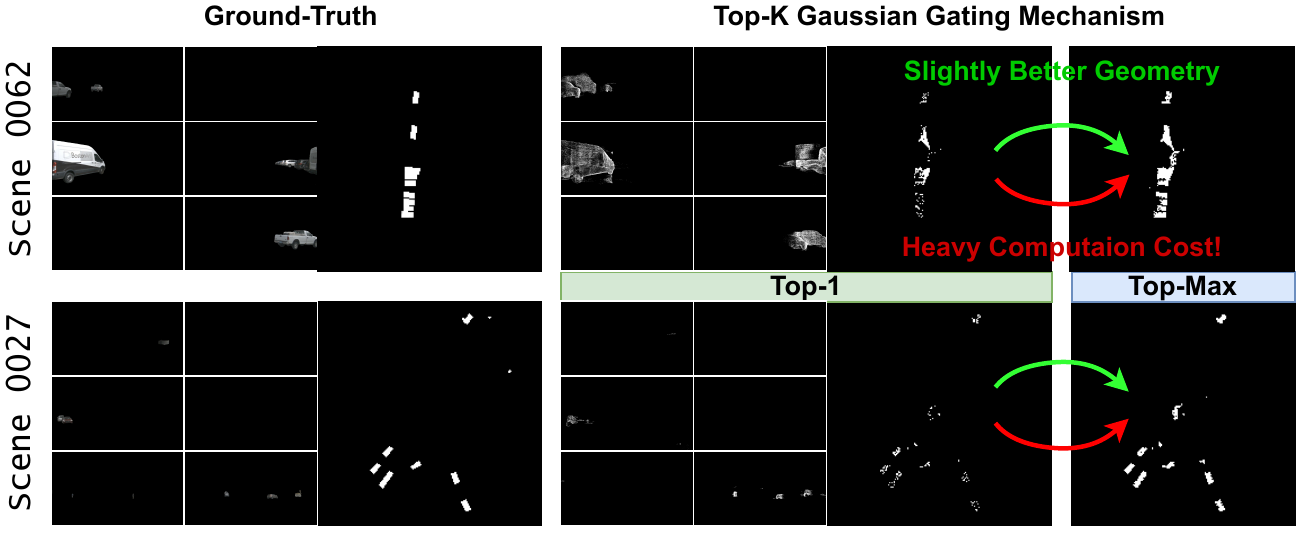}
   \caption{Qualitative assessment of top-$K$ Gaussian gating in image and BEV.}
   \label{fig:sup:top1gate}
\end{figure}

We extend the gating strategy introduced in~\cref{sec:method:c3} to a generalized top-$K$ formulation, specifically investigating the empirical trends as $K$ increases. As shown in~\cref{fig:sup:top1gate}, increasing $K$ toward the dense limit of teacher primitives facilitates a more precise geometric alignment by integrating a broader set of spatial priors. In the visualization, the white dots denote the projected centers of the splatted 3D Gaussians, illustrating the increased structural density as $K$ scales. However, this gain in precision is accompanied by a linear increase in the number of 3D Gaussians and a consequent rise in computational demands, escalating the model's training overhead and memory footprint during the distillation phase.

Our empirical findings demonstrate that, despite the increased precision at higher $K$ values, the top-1 configuration provides a sufficient foundation for robust 3D grounding. As observed in the top-1 visualizations, the Gaussian with the maximal contribution weight already encapsulates the fundamental geometric and semantic essence of the target scene. Since the primary primitive dominates the unprojection signal, the incremental structural gains observed when $K$ approaches its maximum follow a trend of diminishing returns. Consequently, we justify $K=1$ as the optimal default design; it strikes an ideal balance by capturing the core 3D geometry through a clear one-to-one correspondence, thereby avoiding the redundant and high-cost overhead of multi-Gaussian aggregation.

\begin{table}[tb]
    \centering
    \caption{\textbf{3D detection results on the nuScenes benchmark.} The evaluation follows a rigorous OV protocol where all supervision for novel classes is removed.}
    \label{tab:sup:3dod}
    
    \resizebox{0.4\textwidth}{!}{%
        \begin{tabular}{lcc}
            \toprule
            \textbf{Method} & \textbf{NDS} & \textbf{mAP} \\
            \midrule
            BEVDet~\cite{bevdet}           & 27.1 & 21.9 \\
            BEVFormer~\cite{bevformer}     & 27.5 & 19.5 \\
            GaussianLSS~\cite{gaussianlss} & 26.4 & 20.6 \\
            \rowcolor{gold}
            \textbf{OVBEVSeg (Ours)}       & \textbf{30.8} & \textbf{24.1} \\
            \bottomrule
        \end{tabular}
    }
\end{table}

\subsection{Open-Vocabulary 3D Object Detection}
To evaluate the broader zero-shot generalization capabilities of OVBEVSeg, we extend our evaluation beyond instance-level BEV segmentation to OV 3D object detection. Following standard OV protocols~\cite{ov,ovsurvey}, we establish a stringent unseen scenario by excluding all GT annotations for the three novel classes (\ie, ``truck,'' ``bus,'' and ``motorcycle''). We integrate a detection head directly onto the learned BEV features, following the architecture of BEVFormer~\cite{bevformer}. Performance is quantified using the mean average precision (mAP) and the nuScenes detection score (NDS). As demonstrated in~\cref{tab:sup:3dod}, our framework not only excels in open-vocabulary BEV segmentation but also demonstrates robust cross-task extensibility, effectively handling complex 3D detection in the absence of direct supervision for novel categories.

\section{Additional Qualitative Results}
In this section, we provide comprehensive visual evidence to demonstrate the efficacy of our offline framework. Through a series of top-down and perspective visualizations, we illustrate how our method achieves superior spatial consistency and high-fidelity 3D scene reconstruction.

\begin{figure}[tb]
  \centering
   \includegraphics[width=\linewidth]{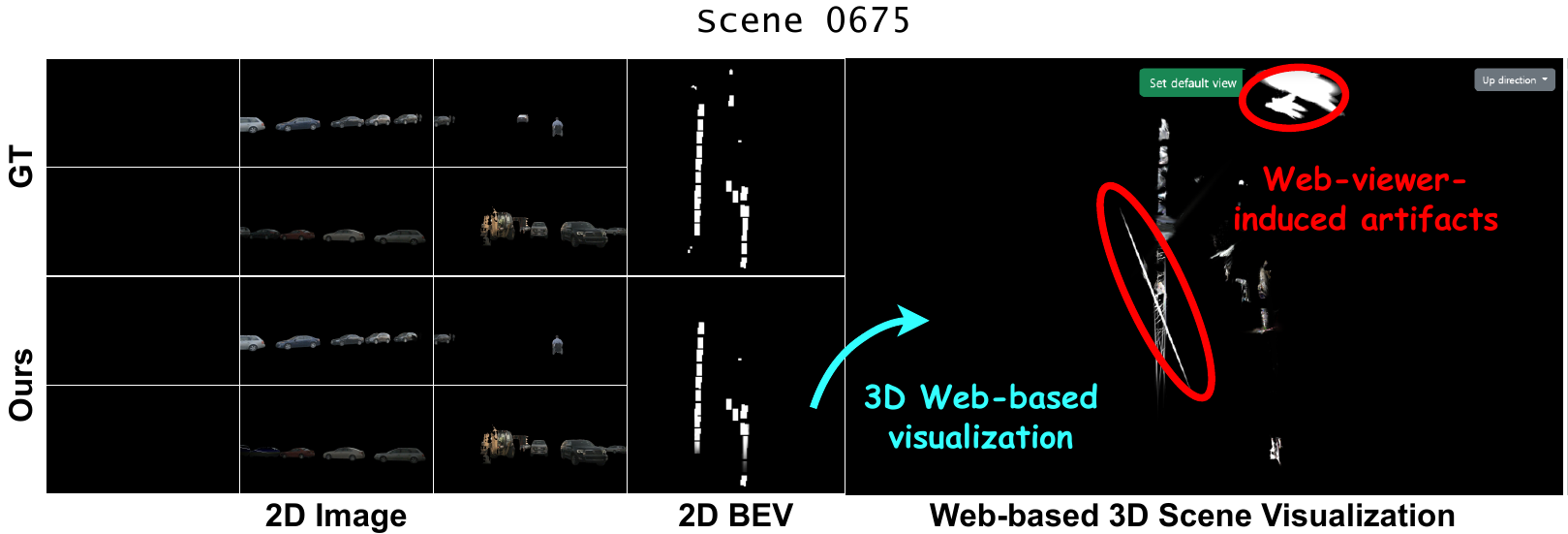}
   \caption{\textbf{Qualitative results of cross-modal consistency in 2D--BEV--3D reconstruction via BAGS.} The 3D renderings were captured using a web-based visualization tool, which may introduce viewer-induced ghosting or smearing artifacts.}
   \label{fig:sup:3drecon2}
\end{figure}

\subsection{3D Scene Reconstruction  and Geometric Fidelity}
\subsubsection{3D scene visualization.}
\begin{figure}[tb]
  \centering
  \includegraphics[width=\linewidth,height=\textheight,keepaspectratio]{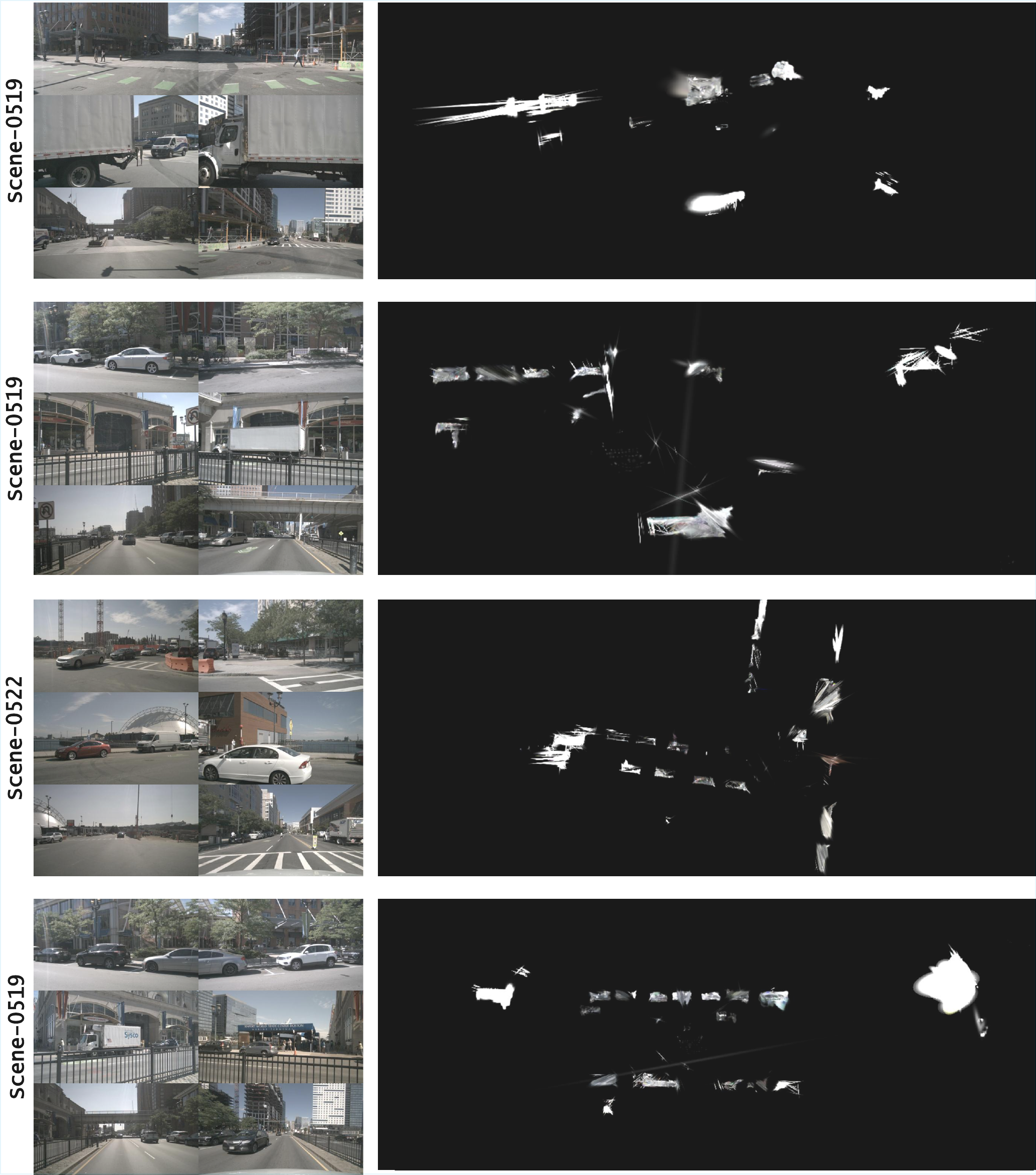}
  \caption{\textbf{Qualitative assessment of 3D scene reconstruction via BAGS.} These novel views were captured using an external web-based visualization tool, which may introduce minor viewer-induced ghosting or smearing artifacts.}
  \label{fig:sup:3drecon}
\end{figure}

\cref{fig:sup:3drecon2,fig:sup:3drecon} present qualitative joint 2D--3D--BEV reconstructions via the BAGS framework; 3D renderings are shown in an approximately orthographic projection. These 3D novel views were captured by exporting the optimized 3D Gaussians into a third-party web-based visualization tool. We observe that the resulting 3D renderings may exhibit minor viewer-induced 3D ghosting artifacts, as highlighted in~\cref{fig:sup:3drecon2}; however, these are purely visualization-side artifacts and are unrelated to our proposed contributions. In practice, direct inspection of the optimized primitives confirms that no 3D Gaussians reside within the proximity of these artifacted regions in the source data. These visualizations are provided for reference to facilitate a more intuitive understanding of the overall 3D scene reconstruction.

In short, these qualitative results demonstrate that our approach maintains high-fidelity image reconstruction quality---ensuring multi-view consistency---while effectively resolving the inherent spatial ambiguities in the BEV plane often overlooked by existing methods. For instance, in ``Scene-0522,'' the top-down perspective clearly preserves fine-grained object-level geometry. This structural integrity facilitates the generation of more precise object-centric BEV representations when projecting the 3D Gaussian primitives onto the BEV grid. Collectively, our method is the first to establish that BEV-level supervision signals can serve as a potent geometric prior, enabling the optimization of 3D Gaussians with superior spatial accuracy and yielding more faithful 3D scene representations.

\clearpage

\subsubsection{2D--depth--BEV visualization.}
\begin{figure}[tb]
  \centering
  \includegraphics[width=\linewidth,height=\textheight,keepaspectratio]{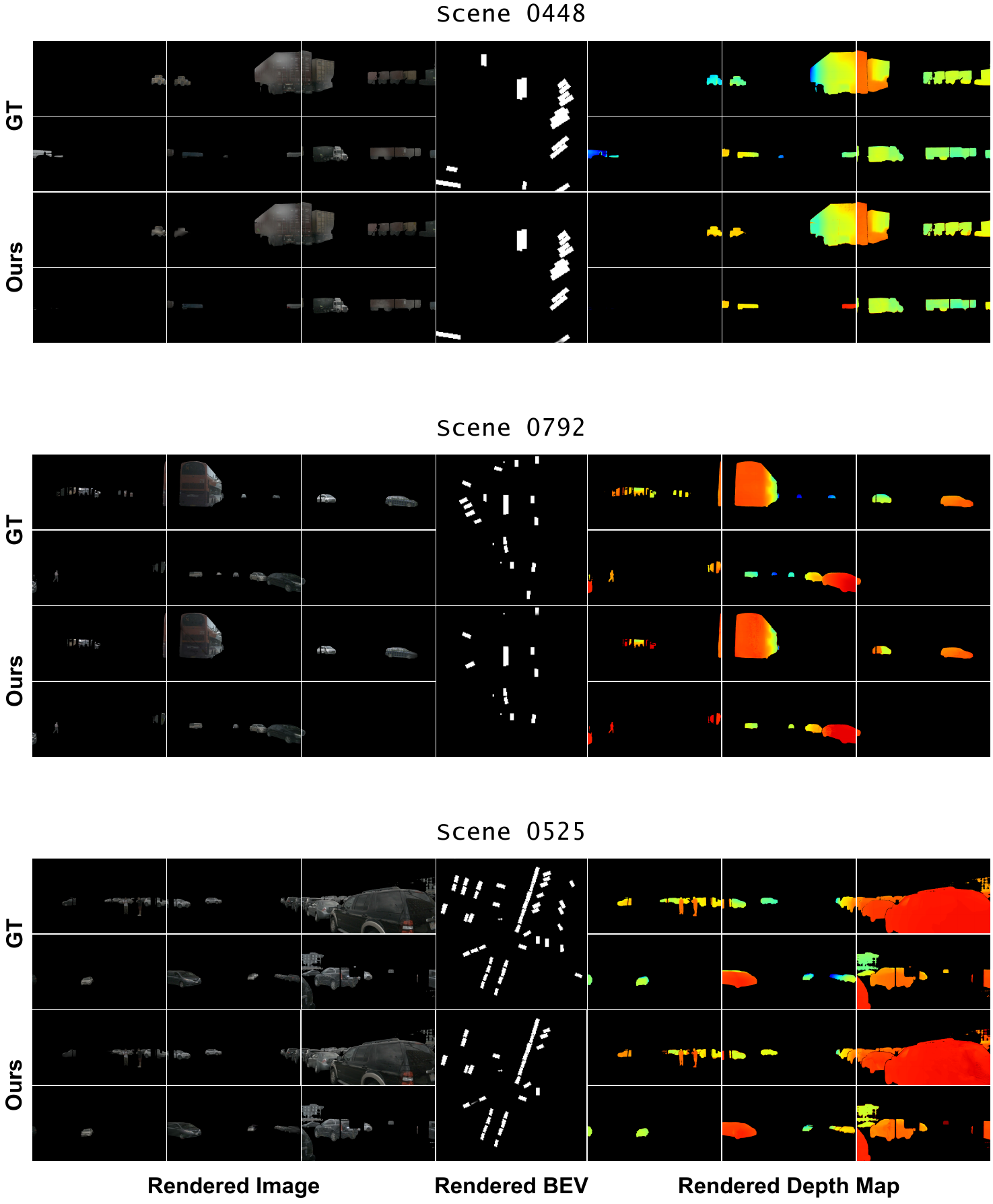}
  \caption{Qualitative validation of 2D--depth--BEV scene reconstruction via BAGS.}
  \label{fig:sup:2dbev}
\end{figure}

\Cref{fig:sup:2dbev} illustrates that our approach yields high-fidelity image and depth reconstructions while effectively mitigating the spatial ambiguities within the BEV plane that typically remain unaddressed by vanilla 3DGS~\cite{3dgs}. Concretely, plain 3DGS exhibits significant structural collapse in the BEV domain, losing the necessary spatial consistency for reliable scene reconstruction. In stark contrast, our top-down perspective faithfully preserves fine-grained object-level geometry, with the corresponding depth maps accurately recovering the underlying spatial layout. This structural integrity facilitates the generation of more precise object-centric BEV representations when projecting the 3D Gaussian primitives onto the BEV grid. Collectively, our method is the first to establish that BEV-level supervision signals can serve as a potent geometric prior, especially under sparse on-vehicle viewpoints.

\section{Baseline Models}
This section details the foundational architectures integrated into our framework, which provide the basis for both OV perception and geometric distillation.

\paragraph{GaussianLSS.}
GaussianLSS~\cite{gaussianlss} is an uncertainty-aware BEV perception framework that reformulates the Lift-Splat-Shoot paradigm~\cite{lss}. It estimates per-pixel depth distributions and lifts them into a collection of 3D Gaussian primitives, which are subsequently rasterized into BEV features via differentiable Gaussian splatting. This approach achieves a superior balance between accuracy and inference throughput among unprojection-based methods. In our framework, GaussianLSS serves as the feed-forward 3DGS backbone, upon which we integrate our OV segmentation head and apply our proposed PBL and BAGD strategies.

\paragraph{Per-scene optimization (\ie, offline 3DGS).}
Per-scene optimization facilitates high-fidelity scene reconstruction by representing complex environments through anisotropic 3D Gaussian primitives. Each primitive is characterized by its spatial extent and appearance, which are iteratively refined to ensure multi-view consistency. The parameters of these primitives---including position, covariance, opacity, and color---are optimized from multi-view imagery. By utilizing a visibility-aware tiled rasterizer, 3DGS enables the rendering of high-quality radiance fields at real-time frame rates. Building upon this, we develop BAGS to serve as a high-fidelity geometric teacher. BAGS provides the dense supervision signal required to guide the student BEV network during the distillation phase of BAGD.

\paragraph{UNION.}
UNION~\cite{union} is a state-of-the-art unsupervised 3D object detection framework. it generates static and dynamic object proposals from LiDAR point clouds using spatial clustering and self-supervised scene flow estimation. These proposals are grouped into pseudo-classes to train 3D detectors without requiring manual annotations. In our pipeline, we utilize UNION’s class-agnostic 3D bounding boxes as candidate object proposals. These proposals are instrumental in establishing 2D--BEV spatial correspondences, allowing us to generate high-quality BEV pseudo-labels for both base and novel semantic categories.

\clearpage

\section{View Transformations}
In this section, we describe the geometric transformations into both the 2D perspective image plane (\cref{sec:sup:3dproj}) and the BEV space (\cref{sec:sup:bevsplat}).

\subsection{3D Camera Projection}
\label{sec:sup:3dproj}
To establish the geometric relationship between the 3D world and the 2D image plane, we adopt the standard pinhole camera model. A 3D point in the world coordinate system, $\mathbf{P}_w = [x_w, y_w, z_w]^\top$, is first transformed into the camera coordinate system $\mathbf{P}_c = [x_c, y_c, z_c]^\top$ using the camera extrinsic parameters:
\begin{equation}
    \mathbf{P}_c = R \mathbf{P}_w + t,
\end{equation}
where $R \in \mathbb{R}^{3\times3}$ is the rotation matrix and $t \in \mathbb{R}^3$ is the translation vector. The camera-space point $\mathbf{P}_c$ is then projected onto the 2D image plane to obtain the pixel coordinates $\mathbf{p} = [u, v]^\top$. This process is governed by the camera intrinsic matrix $K$:
\begin{equation}
    z_c \begin{bmatrix} u \ v \ 1 \end{bmatrix} = K \mathbf{P}_c =
    \begin{bmatrix}
    f_x & 0 & c_x \\
    0 & f_y & c_y \\
    0 & 0 & 1
    \end{bmatrix}
    \begin{bmatrix} x_c \ y_c \ z_c \end{bmatrix},
\end{equation}
where $f_x, f_y$ represent the focal lengths and $c_x, c_y$ denote the principal point (optical center). After performing perspective division by the depth $z_c$, the final image coordinates are derived as:
\begin{equation}
    u = f_x \frac{x_c}{z_c} + c_x, \quad v = f_y \frac{y_c}{z_c} + c_y.
\end{equation}
This projection framework provides the fundamental mapping required to align 3D spatial features with their corresponding 2D image-plane observations.

\subsection{BEV Splatting: BEV Features via Gaussian Splatting}
\label{sec:sup:bevsplat}
Following the 3DGS framework~\cite{3dgs}, we represent the 3D scene using a collection of Gaussian primitives, each defined as $g_i = (\mu_i, \Sigma_i, F_i, \alpha_i)$. Here, $\mu_i \in \mathbb{R}^3$ denotes the mean position, $\Sigma_i \in \mathbb{R}^{3\times 3}$ is the covariance matrix representing the spatial extent, $F_i$ is the high-dimensional feature vector, and $\alpha_i$ represents the opacity. These primitives are initialized from per-pixel depth distributions estimated by the image encoder and subsequently projected onto the BEV plane to construct the foundational BEV feature map.

Consistent with GaussianLSS~\cite{gaussianlss}, we project these 3D Gaussians by marginalizing the vertical ($z$) axis. A fixed 2D transformation matrix, $S_{\text{BEV}}$, is applied to swap the $x$--$y$ axes and map metric coordinates to the discrete BEV grid:
\begin{equation}
    S_{\text{BEV}} =
    \begin{bmatrix}
        0 & \text{scale}_x \\
        \text{scale}_y & 0
    \end{bmatrix}.
\end{equation}
Let $\Sigma_{i,xy}$ be the $2\times2$ submatrix of $\Sigma_i$ corresponding to the horizontal coordinates:
\begin{equation}
    \Sigma_{i,xy} =
    \begin{bmatrix}
        \Sigma_{11} & \Sigma_{12} \\
        \Sigma_{21} & \Sigma_{22}
    \end{bmatrix}.
\end{equation}
The covariance of the projected 2D Gaussian in the BEV space is derived as:
\begin{equation}
    \Sigma_i^{\text{BEV}} = S_{\text{BEV}} \, \Sigma_{i,xy} \, S_{\text{BEV}}^\top := \begin{bmatrix}
        \Sigma_{22} \, \text{scale}_x^{2} & \Sigma_{21} \, \text{scale}_x \text{scale}_y \\
        \Sigma_{12} \, \text{scale}_x \text{scale}_y & \Sigma_{11} \, \text{scale}_y^{2}
    \end{bmatrix}.
\end{equation}
The mean vector $\mu_i$ is transformed analogously to obtain the grid-aligned mean $\mu_i^{\text{BEV}}$. Finally, the cumulative BEV features for each grid cell are synthesized by splatting these 2D Gaussians onto the grid, where feature vectors $F_i$ are aggregated via density- and opacity-weighted summation.

\end{document}